\documentclass{article}

\usepackage{arxiv}

\usepackage[utf8]{inputenc} 
\usepackage[T1]{fontenc}    
\usepackage{url}            
\usepackage{booktabs}       
\usepackage{amsfonts}       
\usepackage{nicefrac}       
\usepackage{microtype}      

\usepackage[table,xcdraw]{xcolor}
\usepackage{amsfonts}
\usepackage[misc]{ifsym}
\usepackage{amsmath}
\usepackage{booktabs}
\usepackage{amsthm}
\usepackage[colorlinks]{hyperref}
\usepackage{multirow}
\usepackage{mathtools}
\usepackage{thmtools}
\usepackage{enumitem}
\usepackage[ruled,linesnumbered,noend]{algorithm2e}
\usepackage{lipsum,graphicx,subcaption}

\newtheorem{definition}{Definition}
\newtheorem{theorem}{Theorem}
\newtheorem{lemma}{Lemma}

\newtheorem{proposition}{Proposition}

\newtheorem{corollary}{Corollary}

\newtheorem*{proof*}{Proof}
\newtheorem*{proof2*}{Proof of Theorem 2}
\newtheorem*{pfs*}{Proof Sketch}
\usepackage[noend]{algorithmic}

\newcommand{\toolname}{\texttt{GradSAT}}

\usepackage[switch]{lineno}
\usepackage{xcolor}
\title{On Continuous Local BDD-Based Search for Hybrid SAT Solving\thanks{The author list has been sorted alphabetically by last name; this
should not be used to determine the extent of authors’ contributions.}}
\author{Anastasios Kyrillidis,  Moshe Y. Vardi, Zhiwei Zhang \thanks{Corresponding author: Zhiwei Zhang.}\\ 
Rice University, Houston, TX, USA \\
\{anastasios,  vardi, zhiwei\}@rice.edu
}

\begin{document}
\maketitle
 
\setcounter{secnumdepth}{1} 

%

\begin{abstract}
We explore the potential of continuous local search (CLS) in SAT solving by proposing a novel approach for finding a solution of a hybrid system of Boolean constraints. The algorithm is based on CLS combined with belief propagation on binary decision diagrams (BDDs). Our framework accepts all Boolean constraints that admit compact BDDs, including symmetric Boolean constraints and small-coefficient pseudo-Boolean constraints as interesting families. We propose a novel algorithm for efficiently computing the gradient needed by CLS. We study the capabilities and limitations of our versatile CLS solver, \toolname, by applying it on many benchmark instances.  The experimental results indicate that \toolname{} can be a useful addition to the portfolio of existing SAT and MaxSAT solvers for solving Boolean satisfiability and optimization problems. 
\end{abstract}
\section{Introduction}
Constraint-satisfaction problems (CSPs) are fundamental in mathematics, physics, and computer science.  The Boolean SATisfiability problem (SAT) is a paradigmatic class of CSPs, where each variable takes value from the binary set $\{\texttt{True}, \texttt{False}\}$.  Solving SAT efficiently is of utmost significance in computer science, both from a theoretical and a practical perspective. As a special case of SAT, formulas in conjunctive normal form (CNFs) are a conjunction (\texttt{and}-ing) of disjunctions (\texttt{or}-ing) of literals. Despite the NP-completeness of CNF-SAT, there has been dramatic progress on the engineering side of CNF-SAT solvers \cite{Vardi14a}.  SAT solvers can be classified into complete and incomplete ones: a complete SAT solver will return a solution if there exists one or prove unsatisfiability if no solution exists,  while an incomplete algorithm is not guaranteed to find a satisfying assignment nor can it prove unsatisfiability.

Most modern complete SAT solvers are based on the the Conflict-Driven Clause Learning (CDCL) algorithm introduced in GRASP \cite{marques1999grasp}, an evolution of the backtracking Davis-Putnam-Logemann-Loveland (DPLL) algorithm \cite{davis1960computing,davis1962machine}. Examples of highly efficient complete SAT solvers include Chaff \cite{chaff_paper}, MiniSat \cite{minisat}, PicoSAT \cite{PicoSAT}, Lingeling \cite{lingeling}, Glucose \cite{glucose},  and machine learning-enhanced MapleSAT \cite{maplesat}. Overall, CDCL-based SAT solvers constitute a huge success for SAT problems, and have been dominating in the research of SAT solving.

Discrete local search (DLS) methods are used in incomplete SAT solvers. The number of unsatisfied constraints is often regarded as the objective function. DLS algorithms include greedy local search (GSAT) \cite{GSAT} and random walk GSAT (WSAT) \cite{walksat}. 
Several efficient variants of GSAT and WSAT have been proposed, such as NSAT \cite{Nsat}, Novelty+ \cite{novelty}, SAPS \cite{SAPS}, ProbSAT \cite{probSAT}, and CCAnr \cite{ccanr}. While practical DLS solvers could be slower than CDCL  solvers, they are useful for solving  certain classes of instances, e.g., hard random 3-CNF and MaxSAT \cite{GSAT}. 

Continuous local search (CLS) algorithms are much less studied in SAT community, compared with CDCL and DLS methods. In \cite{SATInteriorPoint}, the SAT problem is regarded as an integer linear programming (ILP) problem and is solved by interior point method after relaxing the problem to linear programming (LP).  Another work \cite{continuouslocalsearch}  converts SAT into an unconstrained global minimization problem and applies coordinate descent. Those methods, however, are only able to solve relatively easy random CNF instances and fail to provide interesting theoretical guarantees regarding rounding and convergence. 

Non-CNF constraints are playing important roles in  computer science and other engineering areas, \emph{e.g.}, XOR constraints in cryptography \cite{Biclique-Cryptanalysis-of-the-Full-AES}, as well as Pseudo-Boolean constraints and Not-all-equal (NAE) constraints in discrete optimization \cite{Graph-coloring-with-cardinality-constraints,nae-coloring}. 
The combination of different types of constraints enhances the expressive power of Boolean formulas. Nevertheless, compared to that of CNF-SAT solving, efficient SAT solvers that can handle non-CNF constraints are less well studied. 

There are two main approaches to handle hybrid 
Boolean systems, i.e., instances with non-CNF constraints:  1) CNF encoding and 2) extensions of existing SAT solvers. 
For the first approach, different encodings differ in size, the ability to detect inconsistencies by unit propagation (arc consistency), and solution density \cite{encoding-handbook-of-satisfiability}. 
It is generally observed that the running time of SAT solvers relies heavily on the detail of encodings. Finding the best encoding for a solver usually requires considerable testing and comparison \cite{Exploiting-Cardinality-Encodings-in-Parallel-Maximum-Satisfiability}. In addition,  the encodings of different constraints generally do not share extra variables, which increases the number of variables. The second approach often requires different techniques for various types of constraints, e.g.,  cryptoMiniSAT \cite{cmspaper} uses Gaussian Elimination to handle XOR constraints, while Pueblo \cite{Pueblo} leverages cutting plane to take care of pseudo-Boolean constraints. 
Despite the efforts of Dixon et a.~\cite{zap} to develop a general satisfiability framework based on group theory, the study of a uniform approach for solving hybrid constraints is far from mature.

Recently, an algebraic framework named FourierSAT \cite{fouriersat} attempted to address the issues with handling non-CNF constraints, bringing CLS methods back to the view of SAT community. As an algorithmic application of Walsh-Fourier analysis of Boolean functions, FourierSAT transforms Boolean constraints into multilinear polynomials, which are then used to construct the objective function, where a solution is found by applying vector-wise gradient-based local search in the real cube $[-1,1]^n$. In this framework, different constraints are handled uniformly in the sense that they are all treated as polynomials. Compared with the earlier CLS algorithm \cite{continuouslocalsearch}, FourierSAT is able to handle more types of constraints than just random CNF formulas. Moreover, FourierSAT provides interesting theoretical properties of rounding as well as global convergence speed thanks to a better-behaved objective function. In addition, taking advantage of the infrastructure for training neural networks, including software libraries and hardware, to solve discrete problems has become a hot topic recently \cite{modelCountingbyTensor}.  The gradient descent-based framework of FourierSAT provides the  SAT community a possible way of benefiting from machine-learning methods and tools.
    
Though FourierSAT does bring attention back to CLS methods, it still has some limitations. First, as a CLS approach, FourierSAT suffers from inefficient gradient computation. Specifically, computing the gradient of $n$ coordinates can take $O(n)$ function evaluations. Second, the types of constraints accepted by FourierSAT are still somehow limited---only constraints with closed-form Walsh-Fourier expansions can be handled. Third, FourierSAT uses random restart when getting stuck in a local optimum, and the information of the previous local optima is not leveraged. 

\medskip
\noindent \textbf{Contributions.} 
We point out here that the core of CLS methods for SAT solving is the ability of efficiently evaluating and differentiating (computing the gradient) the objective, independent of the representations we use for Boolean constraints. We show that the evaluation and differentiation problem in the framework of FourierSAT are equivalent with weighted model counting and circuit-output probability. Model counting and circuit-output probability are hard computational problems for general Boolean functions \cite{countingIsHard}. Nevertheless, if a Boolean function can be compactly represented by certain structures, e.g., Binary Decision Diagram (BDD) \cite{BDD}, then those tasks can be done efficiently. As a widely-used representation of Boolean constraints, BDD has applications in synthesis, verification and counting. Thus practical packages for handling BDDs, e.g., CUDD \cite{cudd} have been developed. Such packages provide sub-function sharing between individual BDDs so that a set of BDDs can be stored compactly. 
    
We propose a novel algorithm for computing the gradient of the objective based on \emph{belief propagation} on BDDs. We prove that by our algorithm, the complexity of computing the gradient is linear in terms of the size of shared BDDs. In fact, computing gradient for $n$ coordinates is as cheap as evaluating the objective function. 
    
We then extend FourierSAT to a more versatile incomplete SAT solver, which we call \toolname{} by using a more general data structure, BDDs, rather than the Walsh-Fourier expansions.  As interesting families of Boolean functions that have compact BDDs, pseudo-Boolean constraints and symmetric Boolean constraints, including CNF, XOR and cardinality constraints, have abundant applications.  We  enhance the performance of \toolname{} in a variety of ways. We borrow the idea of adaptive weighting form DLS  \cite{breakOut} and  Discrete Lagrange Method \cite{DLM} to make use of solving history and search the continuous space more systematically.  We also take advantage of a portfolio of continuous optimizers and  parallelization.
     
In the experimental section, we aim to demonstrate the capabilities and limitations of CLS approaches represented by our tool, \toolname. On one hand, our method owns nice versatility and can outperform CDCL and DLS solvers on certain problems, showing the power of CLS and hybrid Boolean encodings instead of pure CNF. On the other hand, we observe that on certain benchmarks, e.g., formulas with XORs, CDCL solvers still offer better performance, revealing potential limitations of CLS methods for handling instances with solution-space shattering. We conclude that CLS methods can be a useful complement to strengthen the SAT-solver portfolio. In addition, since local search-based SAT solvers can be naturally used to solve MaxSAT problems \cite{Incomplete-algorithms}, we apply \toolname{} to many MaxSAT instances and obtain  encouraging results.

\section{Preliminaries}
\label{section:pre}
\subsection{Boolean Formulas, Constraints and BDDs}
	Let $X=(x_1,...,x_n)$ be a sequence of $n$ Boolean variables. A Boolean function $f(x)$ is a mapping from a Boolean vector $\{\texttt{True,False}\}^n$ to $\{\texttt{True},\texttt{False}\}$.
	In this paper, we define a Boolean function by $f: \{\pm 1\}^n \to \{0,1\}$, where for variables, $-1$  stands for \texttt{True} and $+1$ for \texttt{False} and for the value, $1$ stands for \texttt{True} and $0$ for \texttt{False}. 
	A vector $a\in\{-1,1\}^n$ is called an assignment. 
	A literal $l_i$ is either a variable $x_i$ or its negation $\neg x_i$. In the rest of this paper, we focus on the case where  $f=c_1\wedge c_2\wedge \dots \wedge c_m$. In other words, $f$ is the logical-AND of $m$ Boolean functions, where each $c_i$ is called a constraint. 
	\begin{itemize}
	    \item \emph{Symmetric Boolean constraints:} Assume every literal appears at most once in a constraint. Let $c(l_1\cdots l_{n_c})$ be  a constraint determined by literals $(l_1\cdots l_{n_c})$. $c$ is called \textbf{symmetric} if  $c(l_1\cdots l_{n_c})\equiv c(\delta(l_1),\cdots,\delta(l_{n_c}))$ holds for every permutation $\delta$ on $(l_1\cdots l_{n_c})$.
	   Disjunctive clauses are symmetric.  For example, $(x_1\vee \neg x_2 \vee x_3)\equiv (\neg x_2\vee x_3 \vee x_1)$. Other interesting classes of symmetric constraints include cardinality constraints (CARD), e.g., $x_1+x_2+x_3\ge 2$; XOR,  e.g., $x_1\oplus x_2\oplus x_3$; 
	   Not-all-equal constraints (NAE), which are satisfied when not all the variables have the same value.
	 \item Pseudo-Boolean constraints (PB), e.g,  $3x_1+5\neg x_2-6x_3\ge 2$, where the coefficients are limited to be integers.
	\end{itemize}
	
	 Let the set of constraints of $f$ be $C_f$. Let $m=|C_f|$ and $n$ be the number of variables of $f$. A solution of $f$ is an assignment that satisfies all the constraints in $C_f$.
	 
A Binary Decision Diagram, or BDD, is a data structure that represents a Boolean function as a directed acyclic
graph $B$ \cite{BDD}.  We use $B.V$, $B.E$ and $S(B)=|B.V|+|B.E|$ to denote the node set, edge set and size of $B$. For a node $v\in B.V$, we use $v.T$, $v.F$ and $i_v$ to denote the \texttt{True}, \texttt{False} child and the variable index associated with $v$, respectively.  We use $B.one$ and $B.zero$ to denote the terminal nodes with value $1$ and $0$, respectively. We also use multi-rooted BDDs (MRBDDs) for representing a group of Boolean functions, in which the BDD can have multiple entries, stored in $B.entry:C_f\to B.V$. Each entry of the MRBDD is the root of the single-rooted BDD of one Boolean function. 

	\subsection{Walsh-Fourier Expansion of a Boolean Function}
	
	Walsh-Fourier Transform is a method for transforming a Boolean function into a multilinear polynomial. The following theorem states that every function defined on a Boolean hyper-cube has an equivalent polynomial representation.  

	\begin{theorem} \label{FourierTransformation}
	{\rm (\cite{O'Donnell:2014:ABF:2683783}, Walsh-Fourier Expansion)} Given a function $f: \{\pm 1\}^n \to \mathbb{R}$, there is a unique way of expressing $f$ as a multilinear polynomial, with at most $2^n$ terms, where each term corresponds to one subset of $[n]$, according to:
	
	$$	f(x) = \sum_{S\subseteq [n]} \left( \widehat{f}(S) \cdot \prod_{i\in S}x_i \right),
		$$
		where $\widehat{f}(S)\in \mathbb{R}$ is called Walsh-Fourier coefficient, given $S$, and computed as:
		\begin{small}
		\begin{equation}\nonumber
		    \begin{split}
		        \widehat{f}(S) = \underset{x\sim \{\pm 1\}^n}{\mathbb{E}} \left[f(x) \cdot \prod_{i\in S}x_i \right] 
		        = \frac{1}{2^n} \!\!\! \sum_{x\in \{\pm 1\}^n} \left(f(x) \cdot \prod_{i\in S}x_i\right),
		    \end{split}
		\end{equation}
		\end{small}
		where $x\sim \{\pm 1\}^n$ indicates $x$ is uniformly sampled from $\{\pm 1\}^n$.
		The polynomial is called the \textbf{Walsh-Fourier expansion} of $f$. 
	\end{theorem}
	
	\begin{table}
		\centering
		\begin{tabular}{c c c}
			\toprule 
			$c$ & & $\texttt{WFE}_c$ \\
			\cmidrule{1-1} \cmidrule{3-3}
			$x_1\vee x_2$ & & $ \frac{3}{4}-\frac{1}{4}x_1-\frac{1}{4}x_2-\frac{1}{4}x_1x_2$ \vspace{0.2cm}   \\ 
			$x_1\oplus x_2\oplus x_3$ & & $\frac{1}{2}-\frac{1}{2}x_1x_2x_3$ \vspace{0.2cm} \\ 
			$\text{NAE}(x_1,x_2,x_3)$ & & $\frac{3}{4}-\frac{1}{4}x_1x_2-\frac{1}{4}x_2x_3-\frac{1}{4}x_1x_3$  \\ 
			\bottomrule
		\end{tabular} 
		\caption{Examples of Walsh-Fourier Expansions}
		\label{ex_FE}
	\end{table}
	
	 Given a formula $f$, for each constraint $c$ of $f$,  we use $\texttt{WFE}_c$ to denote the Walsh-Fourier expansion of $c$. Table \ref{ex_FE} shows some examples of Walsh-Fourier expansions.

\subsection{Weighted Model Counting and Circuit-Output Probability}	
\begin{definition}
{\rm (Weighted model counting)}
Let $f:\{-1,1\}^n\to\{0,1\}$ be a Boolean function over a set $X$ of variables. Let $W: \{-1,1\}^n\to \mathbb{R}$ be an arbitrary function. The weighted model counting of $f$ w.r.t. $W$ is
$$
W(f) = \sum_{a\in \{-1,1\}^n}f(a) \cdot W(a).
$$
\end{definition} 
$W$ is called the weight function. In this work we focus on so-called literal-weight functions, where the weight of an assignment can be expressed as the product of weights associated with all satisfied literals:
$
W(a) = \prod_{a_i=-1} W^T(x_i)\cdot \prod_{a_i=1 }W^F(x_i)
$
for some literal weight functions $W^T,W^F:X\to \mathbb{R}$.

When the literal weight functions satisfy $W^T(x_i)+W^F(x_i)=1$ and $W^T(x_i), W^F(x_i)\ge 0$ for each variable $x_i$, the literal weighted model counting problem reduces to the \textbf{circuit-output probability problem}.
\begin{definition}{\rm (Circuit-output probability)}
Let $f$ be a Boolean function over a set $X$ of variables. Let $P: X\to [0,1]$ be the variable input probability. The circuit-output-probability problem of $f$ w.r.t. $P$, denoted by $\texttt{COP}(P,f)$ is the probability of $f$ outputting 1 (\texttt{True}) given the value of each variable independently sampled from the binary distribution with probability $\mathbb{P}[x_i=-1]=P(x_i)$, i.e., 
\begin{small}
$$
\texttt{COP}(P,f) = \sum_{a\in \{\pm 1\}^{|X|}}f(a) \cdot \prod_{a_i=-1} P(x_i)\cdot \prod_{a_i=1}(1-P(x_i))
$$\end{small}
\end{definition}

\section{Theoretical Framework}
In this section, we first recap a previous CLS framework, FourierSAT and give a weighted adaptation of it. Then we show how our new CLS-based method, \toolname{}, can accept an abundant set of different types of constraints and run efficiently due to fast gradient computation. Proofs are delayed to the supplemental material.
	\subsection{Recap of FourierSAT and a Weighted Adaptation}
	
	The basic framework of FourierSAT \cite{fouriersat} is as follows. Given a formula $f$, the \textbf{objective function} w.r.t $f$, denoted by $F_{f}:[-1,1]^n\to \mathbb{R}^+$, is the sum of Walsh-Fourier expansions of $f$'s constraints, i.e., 
	$F_{f}=\sum_{c\in C_f} \texttt{WFE}_c,$
	where $\texttt{WFE}_c$ is the Walsh-Fourier expansion of constraint $c$.
		The following theorem reduces SAT to a multivariate optimization problem over $[-1,1]^n$. 
	\begin{theorem}
	{\rm\cite{fouriersat}}
	\footnote{In \cite{fouriersat} the problem is, in fact, to minimize instead of  maximize, due to the difference of the definition of Boolean functions, which, however, does not bring about an essential differences.}
		\label{red} A Boolean formula $f$ is satisfiable if and only if $\mathop{\max}\limits_{x\in [-1,1]^n}F_{f}(x)=|C_f|.$
	\end{theorem}	
Theorem \ref{red} suggests that we can search \emph{inside} of the Boolean cube instead of only on vertices. 
Then a continuous optimizer is applied for finding the global maximum of the objective. The objective built by Walsh-Fourier expansions is multilinear, which is better-behaved than higher order polynomials. When getting stuck in a local maximum, FourierSAT simply invokes random restart, as Alg. \ref{algo:FourierSAT} shows.
	
A limitation of Theorem \ref{red} is that  different relative importance of constraints is not taken into account. For example, constraints with low solution density are generally harder to satisfy than those with high solution density. We address this issue by involving a \emph{constraint-weight function} $w:C_f\to \mathbb{R}^+$. We define the new objective  as follows. 
	
	\begin{definition} {\rm{(Objective)}} The objective function w.r.t. the formula $f$ and constraint-weight function $w$, denoted by $F_{f,w}:[-1,1]^n\to \mathbb{R}$ is defined as
		$F_{f,w}=\sum\limits_{c\in C_f}w(c)\cdot \texttt{WFE}_c.$
		\end{definition}
	
We now have the weighted analogue of Theorem \ref{red} as follows.
		\begin{theorem} 
		\label{redw} Given a constraint weight function $w:C_f\to \mathbb{R}^+$, a Boolean formula $f$ is satisfiable if and only if $$\mathop{\max}\limits_{x\in [-1,1]^n}F_{f,w}(x)=\sum\limits_{c\in C_f}w(c).$$
		\label{theo:main_weighted}
	\end{theorem}
	
Based on Theorem \ref{theo:main_weighted}, we can design a CLS  framework similar to Alg. \ref{algo:FourierSAT} to search for a maximum of $F_{f,w}$.


\begin{algorithm}[t!]
    \SetAlgoLined
    \SetKwInOut{Input}{Input}
    \SetKwInOut{Output}{Output}
    \Input{Boolean formula $f$ with constraint set $C_f$}
    \Output{A discrete assignment $x\in \{-1,1\}^n$}
    \vspace{0.1cm}
    \hrule
    \vspace{0.1cm}
    \For{$j=1,\; \dots, \;J$}{
        Randomly sample $x_0$ uniformly from $[-1,1]^n$. \\
        Search for a local maximum $x_j^*$ of $F_{f}$ in $[-1,1]^n$, starting from $x_0$.\\
        \lIf{$F_{f}(x^*_j)=|C_f|$}{\Return $x_j^*$}
    }
    \Return  $x_j^*$ with the highest $F_{f}(x_j^*)$ after $J$ iterations
 \caption{FourierSAT, a CLS-based SAT Solver.}
 \label{algo:FourierSAT}
\end{algorithm}


\subsection{Computing Gradients is Critical for CLS Methods}
After constructing the objective function, a CLS approach has to find a global optimum. Global optimization on non-convex functions 
is NP-hard \cite{non-convex-is-nphard}, so
converging to local optima and checking if any of them is global is more practical, as long as global optima can be identified efficiently. 
Algorithms that aim to find local optima can be categorized into several classes of local-search methods, such as gradient-free algorithms, gradient-based algorithms and Hessian-based algorithms \cite{nocedal2006numerical}. Gradient-free algorithms are suitable for cases where the gradient is hard or impossible to obtain \cite{larson2019derivative,berahas2019theoretical}. They are, however, generally inefficient and often their convergence rates depend on the dimensionality of the problem \cite{gradient-free-is-slow}. 
Hessian-based algorithms are able to escape saddle points but are usually too expensive to use \cite{nocedal2006numerical}. 
Thus in practice, using gradient-based algorithms is the most typical choice, which is the case in FourierSAT. The performance of a specific gradient-based algorithm  relies heavily on how fast the gradient can be computed. Since most computational work of CLS-based methods happens in the continuous optimizer  \cite{continuouslocalsearch}, the efficiency of computing gradients is critical for the performance of CLS solvers. 

\subsection{GradSAT: from Walsh-Fourier Expansions to BDDs}
We start this subsection by analyzing some limitations of FourierSAT. 
First, FourierSAT only accepts symmetric constraints.
 There are, however, useful constraints that cannot be reduced to symmetric ones, e.g., pseudo-Boolean constraints (PBs). Second, in FourierSAT, computing gradient costs $O(n)$ function evaluations \cite{fouriersat}, which can get expensive for constraints with more than a few hundred variables. 

The issues above indicate that there might be better representations of Boolean constraints than Walsh-Fourier expansions. In the rest of this section, we discuss how our new framework, \toolname{}, addresses those issues by choosing Binary Decision Diagrams (BDDs) as the representation to evaluate and differentiate the objective. 

In order to use a new representation of Boolean constraints, we need to have a deeper understanding of the objective function. Theorem \ref{theo:rounding} in below, adapted from Theorem 3 in \cite{fouriersat} indicates the objective value can be interpreted as a measure of the progress of the algorithm.
	\begin{definition}
	{\rm{(Randomized rounding)}} We define the randomized rounding function $\mathcal{R}:[-1,1]^n\to \{-1,1\}^n$ by
	\begin{equation}\nonumber
	\begin{cases}
	\mathbb{P}[\mathcal{R}(a)_i=-1]=\frac{1-a_i}{2} \\
	\mathbb{P}[\mathcal{R}(a)_i=+1]=\frac{1+a_i}{2}
	\end{cases}
		\end{equation}
	for $i\in\{1,\cdots,n\}$,	where $\mathbb{P}$ denotes the probability.
	\label{defi:randomized_function}
	\end{definition}
	
	\begin{definition}
	{\rm{(Vector probability space)}} We define a probability space on Boolean vectors w.r.t. real point $a\in[-1,1]^n$, denoted by $\mathcal{S}_a:\{-1,1\}^n\to [0,1]$ by:
	
$$
\mathcal{S}_a(b) = \mathbb{P}[\mathcal{R}(a)=b]= \prod\limits_{b_i=-1}\frac{1-a_i}{2} \prod\limits_{b_i=1}\frac{1+a_i}{2},$$
	for $b\in\{-1,1\}^n$, 
	with respect to the randomized-rounding function $\mathcal{R}$.
	We use $b\sim S_a$ to denote that $b\in\{-1,1\}^n$ is sampled from the probability space $\mathcal{S}_a$.
	\label{defi:vector_space}
	\end{definition}

	\begin{theorem}
	\label{theo:rounding}
 Given a formula $f$ and constraint weight function $w:C_f\to \mathbb{R}^+$, for a real point $a\in [-1,1]^n$, we have
    	  $$  F_{f,w}(a) = \mathop{\mathbb{E}}\limits_{b\sim \mathcal{S}_a}[F_{f,w}(b)],$$
	    \label{equa:expectation}
where $\mathop{\mathbb{E}}$ denotes expectation.
	\end{theorem}
	
For a discrete point $b\in \{-1,1\}^n$, $F_{f,w}(b)$ is the sum of weights of all constraints satisfied by $b$. Therefore, Theorem \ref{theo:rounding} reveals that for a real point $a$, the value $F_{f,w}(a)$ is in fact the expected solution quality at $a$. Thus, a CLS approach can make progress in the sense of expectation as long as it  increases the objective value. Compared with DLS solvers and coordinate descent-based CLS solvers that only flip one bit per iteration, a CLS method is able to "flip" multiple variables, though probably by a small amount, to make progress. 

We now interpret Theorem \ref{theo:rounding} from the perspective of circuit-output probability.
\begin{corollary} With $f,w,a$ as in Theorem \ref{theo:rounding}, we have
	\begin{equation}\nonumber
	    F_{f,w}(a) = \mathop{\mathbb{E}}\limits_{b\sim \mathcal{S}_a}[F_{f,w}(b)]=\sum_{c\in C_f}w(c)\cdot \texttt{COP}(P_a,c),
	    \label{equa:cop}
	    \end{equation}
	   where $P_a:X\to\mathbb{R}$ is the variable input probability function defined by $P_a(x_i)=\frac{1-a_i}{2}$ for all $i\in\{1,\cdots,n\}$. 
	   \label{coro:cop}
\end{corollary}
Corollary \ref{equa:cop} states that to evaluate the objective function, it suffices to  compute the circuit-output probability of each constraint. Calculating the circuit-output probability, as a special case of literal weighted model counting, is \#P-hard for general representations such as CNF formulas  \cite{countingIsHard}. Nevertheless, if a Boolean function is represented by a BDD $B$ of size $S(B)$, then the circuit-output probability on it can be solved in time $O(S(B))$ by a probability assignment algorithm  \cite{probabilityAssignment} shown in Alg. \ref{algo:probabilityAssignment}.

\begin{lemma}  {\rm\cite{probabilityAssignment}}
Alg. \ref{algo:probabilityAssignment} returns the circuit-output probability $\texttt{COP}(P,f)$, where $P\equiv \frac{1}{2}$ and runs in time $O(S(B))$ given the BDD $B$ of $f$.
\end{lemma}
 
 Alg. \ref{algo:probabilityAssignment} deals with the uniform variable probability, $P(x_i)=\frac{1}{2}$ for $i\in\{1,\dots, n\}$. In the following we adapt it for admitting a general variable input probability function $P_a:X\to [0,1]$.

To use the idea of Alg. \ref{algo:probabilityAssignment}, we need to construct the BDDs for the constraints of  $f$. Note that since different constraints can share same sub-functions, equivalent nodes only need to be stored once. In BDD packages such as CUDD \cite{cudd}, this property is realized by a decision-diagram manager. After combining equivalent nodes, the BDD forest generated from all constraints forms a multi-rooted BDD (MRBDD). The size of the MRBDD can be significantly smaller than the sum of the size of each individual BDD when sub-function sharing frequently appears. Alg. \ref{algo:wmc} shows how to use ideas above to evaluate our objective $F_{f,w}$.

	\begin{algorithm}[t!]
    \SetAlgoLined
    \SetKwInOut{Input}{Input}
    \SetKwInOut{Output}{Output}
    \Input{BDD $B$ of Boolean function $f$;}
    \Output{$\texttt{COP}(P,f)$ with $P\equiv \frac{1}{2}$}
    \vspace{0.1cm}
    \hrule
    \vspace{0.1cm}
    \textbf{Step 1.} Assign probability $1$ for the root node of $B$.\\
    \textbf{Step 2.} If the probability of node $v=M[v]$, assign probability $\frac{1}{2}M[v]$ to the outgoing arcs from $v$.\\
    \textbf{Step 3.} The probability $M[u]$ of node $u$ is the sum of the probabilities of the incoming arcs.\\
    \Return  $M[B.one]$
 \caption{The Probability-Assignment Algorithm} 
 \label{algo:probabilityAssignment}
\end{algorithm}

\begin{theorem}
Alg. \ref{algo:wmc} returns the correct objective value $F_{f,w}(a)$ and runs in time O(S(B)), for the MRBDD $B$ of $f$.
\label{lemma:forward}
\end{theorem}

Alg. \ref{algo:wmc} traverses the MRBDD in a top-down style while the same effect can be achieved by a bottom-up algorithm as well, as shown in Alg. \ref{algo:wmc-bottomup}. Similar ideas of traversing BDDs bottom-up have been used in BDD-based SAT solving \cite{bddsat} and weighted model counting \cite{addmc}.

\begin{theorem}
Alg. \ref{algo:wmc-bottomup} returns the correct objective value $F_{f,w}(a)$ and runs in time O(S(B)), for the MRBDD $B$ of $f$.
\label{lemma:backward}
\end{theorem}

	\begin{algorithm}[t!]
    \SetAlgoLined
    \SetKwInOut{Input}{Input}
    \SetKwInOut{Output}{Output}
    \Input{MRBDD $B$, real point $a\in[-1,1]^n$, constraint weight function $w:C_f\to \mathbb{R}$.}
    \Output{value of $F_{f,w}(a)$}
    \vspace{0.1cm}
    \hrule
    \vspace{0.1cm}
    Let $M_{TD}:B.V\to \mathbb{R}$ be the bottom-up messages and $M_{TD}[u] = 0$ for all the non-entry nodes $u\in B.V$. \\
    \For{each constraint $c\in C_f$}{
    $M_{TD}[B.entry(c)] = w(c)$}
    Sort all nodes of $B$  by topological order to  list $L$.\\
    \For{each node $v \in L$}{
     $ M_{TD}[v.T]+=\frac{1-a[i_u]}{2}\cdot M_{TD}[v] $\\
     $ M_{TD}[v.F]+=\frac{1+a[i_u]}{2}\cdot M_{TD}[v] $\\
    }
    \Return  $M_{TD}[B.one]$
 \caption{Objective Evaluation (Top-Down)}
 \label{algo:wmc}
\end{algorithm}

	\begin{algorithm}[t!]
    \SetAlgoLined
    \SetKwInOut{InputOutput}{Input, Output}
    \SetKwInOut{Output}{Output}
      \InputOutput{Same with Algorithm \ref{algo:wmc}}
    \vspace{0.1cm}
    \hrule
    \vspace{0.1cm}
    Let $M_{BU}:B.V\to \mathbb{R}$ be the bottom-up messages and $M_{BU}[u] = 0$ for all nodes $u\in B.V$ except $M_{BU}[B.one] = 1$. \\
   Sort all nodes of $B$  by topological order to  list $L$.\\
    \For{each node $v \in L$}{
    \For{each node $u$ such that $u.T=v$}{
     $ M_{BU}[u]+=\frac{1-a[i_u]}{2}\cdot M_{BU}[v] $\\}
     \For{each node $u$ such that $u.F=v$}{
     $ M_{BU}[u]+=\frac{1+a[i_u]}{2}\cdot M_{BU}[v] $\\}
    }
    \Return  $\sum\limits_{c\in C_f}(M_{BU}[B.entry(c)]\cdot w(c))$
 \caption{Objective Evaluation  (Bottom-Up)} 
 \label{algo:wmc-bottomup}
\end{algorithm}

So far we have demonstrated how to evaluate the objective by BDDs. In the rest of this section we will show the combination of the bottle-up and top-down algorithms inspires an efficient algorithm for  gradient computation.

\subsection{Gradient is as Cheap as Evaluation in GradSAT}
In this subsection, we propose a fast algorithm for computing the gradient, based on message passing on BDDs. The algorithm runs in $O(S(B))$, given $B$ the MRBDD of formula $f$. Roughly speaking, computing the gradient is as cheap as evaluating the objective function.

 The basic idea is to traverse the MRBDD twice, once top-down and once bottom-up. Then, enough information is generated for applying the differentiate operation on each BDD node. The gradient of a specific variable, say $x_i$, is obtained  from all nodes associated with $x_i$. This idea is borrowed from belief-propagation methods on graph model in probabilistic inference \cite{PEARL1988143,beliefPropogation}. Compared with FourierSAT, where the computation of the gradient for different variables has no overlap, Alg.~\ref{algo:grad} exploits the shared work between different variables.
	
	\begin{algorithm}[t!]
    \SetAlgoLined
    \SetKwInOut{Input}{Input}
    \SetKwInOut{Output}{Output}
     \Input{Same with Algorithm \ref{algo:wmc}}
    \Output{Gradient vector $g\in \mathbb{R}^n$ at point $a$}
    \vspace{0.1cm}
    \hrule
    \vspace{0.1cm}
    Let $g$ be a zero vector with length $n$.\\
    Run Algorithm \ref{algo:wmc} with $(B,a,w)$\\
    Run Algorithm \ref{algo:wmc-bottomup} with $(B,a,w)$\\
    Collect $M_{TD},M_{BU}:B.V\to \mathbb{R}$ from Alg. \ref{algo:wmc} and \ref{algo:wmc-bottomup}\\
    \For{each non-terminal node $v\in B.V$}{
      $g[i_v]+= M_{TD}[v]\cdot(M_{BU}[v.F]-M_{BU}[v.T])$
     }
    \Return  $g$
 \caption{Efficient Gradient Computation}
 \label{algo:grad}
\end{algorithm}

\begin{theorem}
Alg. \ref{algo:grad} returns the gradient $g\in \mathbb{R}^n$ of $F_{f,w}$ at real point $a\in[-1,1]^n$ correctly and runs in time $O(S(B))$, given the MRBDD $B$ of $f$.
\footnote{In fact, line 5-7 of Algorithm \ref{algo:grad} can be integrated to Algorithm \ref{algo:wmc-bottomup} once the top-down traverse is done. Thus computing the gradient costs traversing the MRBDD twice instead of three times. We use this trick in our implementation.}
\label{theo:gradient_is_fast}
\end{theorem}

Roughly speaking, $M_{TD}[v]$ (Top-Down) has connection with the probability of a BDD node $v$ being reached from the root, while $M_{BU}[v]$ (Bottom-Up) contains information of the circuit-output probability of the sub-function defined by the sub-BDD with root $v$.  Algorithm \ref{algo:grad} can be interpreted as applying the differentiation operation on each node.

\begin{table*}[ht!]
\centering
\begin{small}
\begin{tabular}{c c c c c c c c c c c c c}
	\toprule 
instance type &  & $n$ & & $m$ &  & {max/avg. constraint length} & & FourierSAT/s & & GradSAT/s & & Acceleration ($\times$) \\ 	\cmidrule{1-1} \cmidrule{3-3} \cmidrule{5-5}\cmidrule{7-7}\cmidrule{9-9}\cmidrule{11-11} \cmidrule{13-13}
{3-CNF} & & 500  & &  2000   &  & 3/3  & & {0.187}  &  & {0.012}                        & &{15.58}   \\ 
10-CNF       & & 1000 & & 32000 & & 10/10  & & 18.81 & & 2.89                         & &6.26    \\ 
2-CNF+1 CARD & & 300 &  & 451  & & 300/2.66 & & 0.40 & & 0.086                       & & 4.65    \\
3-CNF+XORs  &  & 150 & & 360 & & 75/15  & & 0.036  & & 0.029 & & 1.24\\
\bottomrule
\end{tabular}
\end{small}
\caption{\small \textbf{Gradient speedup.} FourierSAT/s and GradSAT/s denote the gradient cost in seconds. \texttt{GradSAT} significantly improves the efficiency of computing gradient.}
\label{table:grad_speedup}
\end{table*}

\subsection{The Versatility of GradSAT}
Now that we have an efficient algorithm to compute the  gradient for formulas with compact BDDs, the next natural question to ask is for what types of constraints can we enjoy small size BDDs. Here we list the following results. 
\begin{proposition} {\rm\cite{symmetric-functions-BDD}}
Symmetric Boolean constraints with $n$ variables admits  BDDs with size $O(n^2)$.
\end{proposition}
\begin{proposition}{\rm\cite{minisatplus}}
Pseudo-Boolean constraints with $n$ variables and magnitude of coefficients bounded by $M$ admits BDDs with size $O(n\cdot M)$.
\end{proposition}

Thus \toolname{} can accept symmetric constraints and small-coefficient PBs, since they have reasonable BDD size. 

It is well known that most of the local search-based incomplete SAT solvers can also work
as a solution approach for the MaxSAT problem, by providing the “best found” truth assignment upon termination \cite{Incomplete-algorithms}. Thus we also use \toolname{} as a partial MaxSAT solver to solve Boolean optimization problems.

\section{Implementation Techniques}
\label{sec:algorithm}
		\subsection{Adaptive Constraint Weighting} When dealing with satisfaction problem, the constraint-weight function $w:C_f\to \mathbb{R}^+$ has a great impact on the performance of CLS method. In practice we find that an appropriate weight function can make it much faster for a continuous optimizer to reach a global maximum. How to pre-define a static weight function for a given class of formulas is, however, still more art than science thus far.
		
		In CLS approaches for SAT solving, solving history is not leveraged if random restart is invoked every time after getting stuck in local optima. We are inspired from the Discrete Lagrange Method \cite{DLM} and the ``break-out"  heuristic \cite{breakOut} in DLS to design the following dynamic weighting scheme. The constraint weights are initialized to be the length of the constraint. When a local optimum is reached, we multiply the weights of unsatisfied constraints by $r$ and start from the same point once again. \toolname{} only does random restart when $T$ trials have been made for the same starting point, as shown in Alg. \ref{algo:GradSAT}.\footnote{We tuned $r$ and $T$ by running \toolname{} on random 3-CNF benchmarks with $r\in\{1,1.5,2,2.5,3\}$ and $T\in\{1,2,4,8,16\}$ and choosing $(r,T)$ that solves most cases. We got $r=2$, $T=8$.}
		\subsection{A Portfolio of Optimizers and Parallelization}
The specific optimizer to maximize the objective also influences the performance of a CLS solver. Different optimizers are good at different types of instances. \toolname{} runs a portfolio of four different optimizers, including SLSQP \cite{slsqp2}, MMA \cite{mma} from NLopt \cite{nlopt} and CG \cite{cg}, BFGS \cite{bfgs} from Dlib \cite{dlib09}, on multiple cores and terminates when any of optimizers returns a solution. 

\begin{algorithm}[t!]
    \SetAlgoLined
    \SetKwInOut{Input}{Input}
    \SetKwInOut{Output}{Output}
    \Input{Boolean formula $f$ with constraint set $C_f$;\\
   Hyperparameters $r$ and $T$
    }
    \Output{A discrete assignment $x\in \{-1,1\}^n$}
    \vspace{0.1cm}
    \hrule
    \vspace{0.1cm}
    Build the MRBDD $B$ of $f$.\\
    \For{$j=1,\; \dots, \;J$}{
     Sample $x_0$ uniformly at random from $[-1,1]^n$. \\
       Set $w(c)=length(c)$ for all constraints $c$.\\ 
       \For{$t=1,\; \dots, \;T$}{
        Starting from $x_0$, search for a local maximum $x_{jt}^*$ of $F_{f,w}$ in $[-1,1]^n$ with Alg. \ref{algo:wmc} for function value and Alg. \ref{algo:grad} for gradient.\\
        \lIf{$F_{f,w}(x^*_{jt})=\sum\limits_{c\in C_f}w(c)$}{\Return $x_{jt}^*$}
    {
         \For{$\forall c\in C_f$ that is not satisfied by $x^*_{jt}$}{Set $w(c)=w(c)\cdot r$}}
    }
    }
    \Return  $x_{jt}^*$ with the highest $F_{f,w}(x_{jt}^*)$ 
 \caption{GradSAT}
 \label{algo:GradSAT}
\end{algorithm}

\section{Experimental Results}
\label{sec:exp}
We aim to answer the following research questions:
\\
\textbf{RQ1.} Is the cost for computing gradient reduced by using the BDD-based approach (Algorithm \ref{algo:grad}) ?\\
\textbf{RQ2. }What is the advantages and limitations of \toolname{}  on randomly generated hybrid Boolean formulas? \\
\textbf{RQ3.} Can \toolname{} perform well in solving discrete optimization problems, encoded by  MaxSAT?   \\    
\textbf{RQ4.} Can MRBDD with nodes sharing significantly reduce the total number of nodes?    

\textbf{Solver Competitors.}
 1) CDCL solvers including Glucose,  cryptoMiniSAT (CMS) and ML-enhanced solver MapleSAT;
    2) DLS solvers, including WalkSATlm \cite{Walksat-implementation}, CCAnr and ProbSAT;
    3) PB solvers including CDCL-based MiniSAT+ \cite{minisatplus}, Open-WBO \cite{openwbo}, NAPS \cite{naps} and Conflict-Driven-Pseudo-Boolean-Search-based (CDPB) solver RoundingSAT \cite{roundingsat};
    4) Partial MaxSAT solvers, including Loandra \cite{loandra}, the best solver of MaxSAT Competition 2019 (incomplete track), 
    WalkSAT (v56) and Mixing Method \cite{wang2017mixing}, an SDP-based solver.
    
\textbf{Benchmark 1: Hybrid random formulas consisting of disjunctive clauses, XORs, CARDs and PBs.} We generate four types of satisfiable random hybrid benchmarks: CNF-XORs (360 instances), XORs-1CARD (270 instances), CARDs (360 instances) and PBs (720 instances) with variable number  $n\in\{50,100,150\}$ \footnote{A detailed description of benchmarks generation can be found in the supplemental material.}. When a type of constraint is not accepted by a traditional CNF solver, we use CNF encodings to resolve the issue. Specifically, we let pySAT \cite{pysat} choose the CNF encoding for CARDs and PBs, while using the methods in \cite{xor-encoding} to encode XORs. Time limit is set to be 100s. 
	
		\textbf{Benchmark 2: MaxSAT problems.} We gathered 570 instances from the \texttt{crafted} (198 instances) and \texttt{random} track (377 instances) of MaxSAT Competition 2016-2019. Those instances encode random problems as well as combinatorial problems such as set covering and MaxCut. We set the time limit to 60s. We also used problems from the incomplete track of MaxSAT Competition 2019 as the \texttt{industrial} benchmarks (296 instances).

\textbf{Implementation of GradSAT.} We implement \toolname{} in C++, with code available in the supplemental material. Each experiment is run on an exclusive node in a  Linux cluster with 16-processor cores at 2.63 GHz and 1 GB of RAM per processor. We generated five versions of \toolname{}, including using single-core on each of the four optimizers (SLSQP, MMA, CG, BFGS) and using $16$ cores on the optimizer portfolio, with each optimizer getting $4$ cores.\\
\begin{figure}[ht]
\centering
\includegraphics[width=1.05\columnwidth]{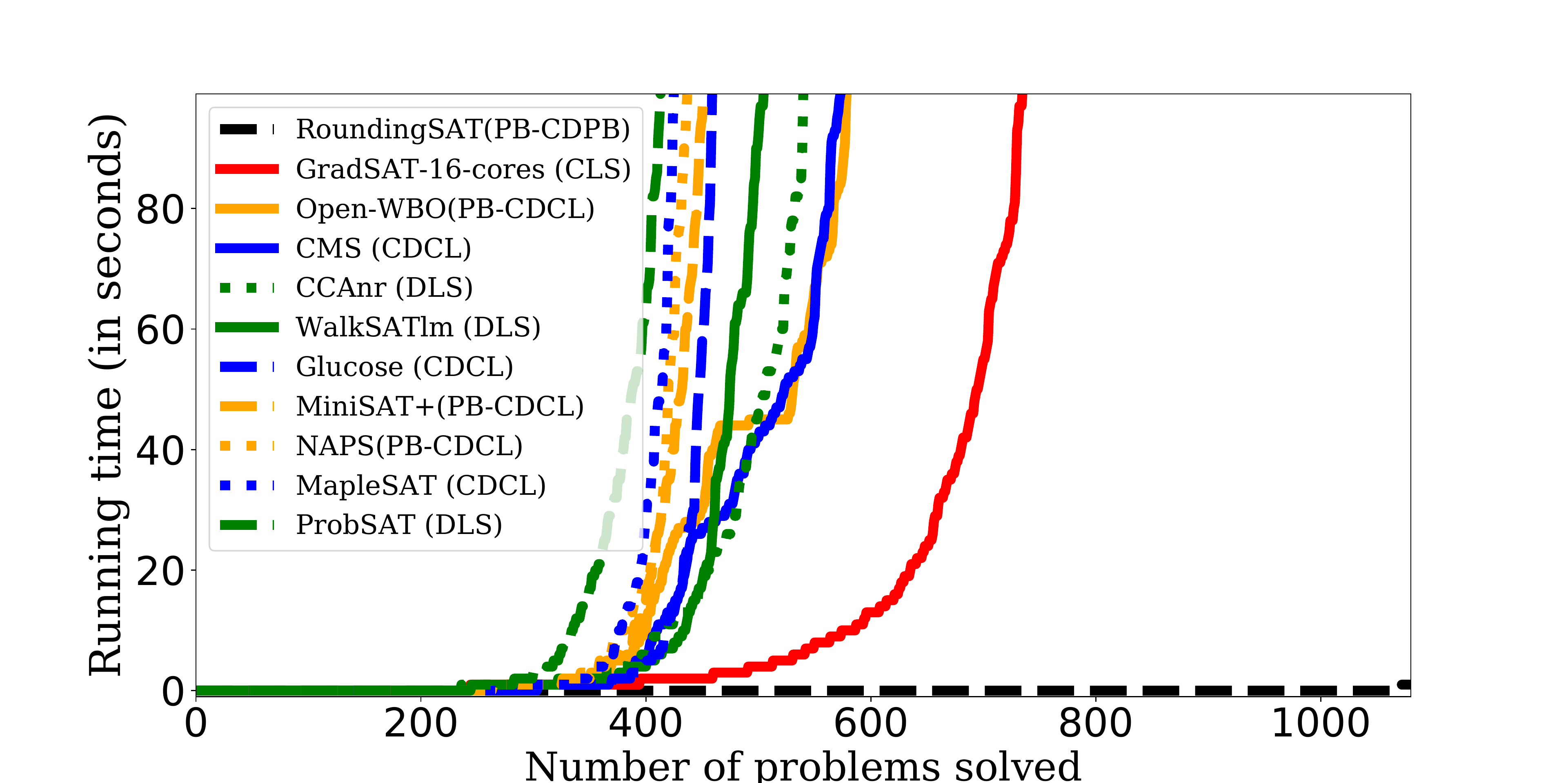}  
\caption{\textbf{Results on random CARDs and PBs.}  \toolname{} is second only to CDPB-based RoundingSAT.} 
\label{exp:rand_card}
\end{figure}

\begin{figure}[ht]
\centering
\includegraphics[width=1.05\columnwidth]{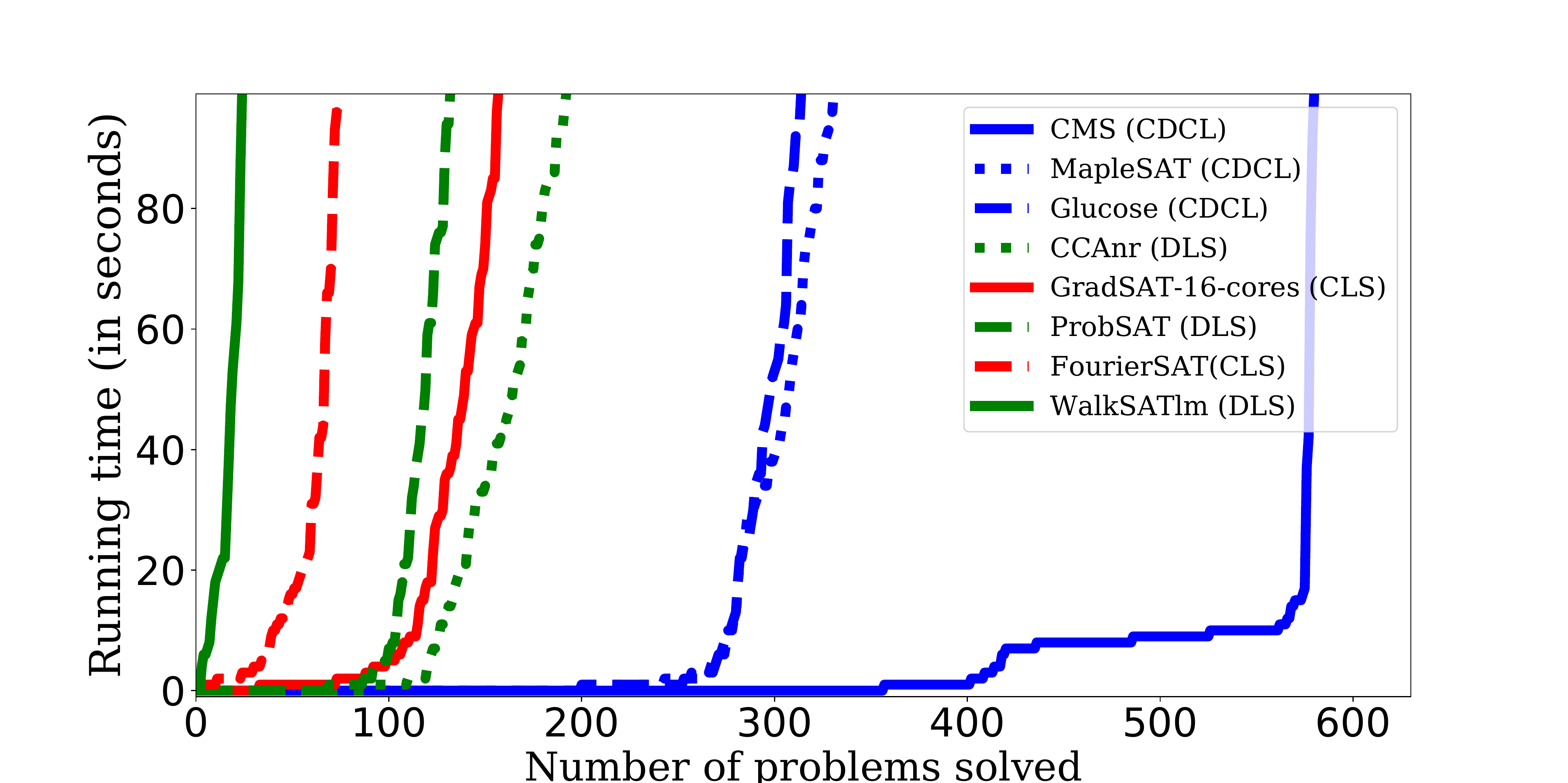}  
\caption{\textbf{Results on random XORs-1CARD and CNF-XORs}. CDCL solvers outperform local-search solvers.} 
\label{exp:rand_xor}
\end{figure}

\begin{table*}[t!]
\centering
\begin{small}
\begin{tabular}{c c c c c c c}
	\toprule 
Benchmarks       &  & Avg. total \# nodes of indi. BDDs & & Avg. \# nodes of MRBDD   & & Avg. redu. ratio ($\times$) \\ 	\cmidrule{1-1} \cmidrule{3-3}\cmidrule{5-5}\cmidrule{7-7}
MaxSAT-Random        &       & 6683 & & 2044 & & 3.22\\ 
MaxSAT-Crafted       &       & 7435 & & 3237 & & 2.51\\ 
MaxSAT-Industrial   &   & 3428810 & & 1548877 & & 2.29\\ 
Hybrid SAT-CARDs       &       & 31714 & & 31454 & & 1.02\\ 
Hybrid SAT-PBs       &       & 602664 & & 602393 & & 1.00\\ 
Hybrid SAT-CNF-XORs      &       & 7254 & & 5600 & & 1.30\\ 
Hybrid SAT-XORs-1CARD      &       & 5976 & & 5794 & & 1.03\\ 
	\bottomrule
\end{tabular}
\caption{\textbf{Nodes sharing in MRBDD}. The average reduction ratio is considerable on MaxSAT instances while negligible on hybrid SAT instances.}
\label{table:nodes-sharing}
\end{small}
\end{table*}

\begin{table}
\centering
\begin{tabular}{c c c c c}
	\toprule 
Methods       &        & Avg. Score &  & \#Win\\ 	\cmidrule{1-1} \cmidrule{3-3} \cmidrule{5-5}
GradSAT-Portfolio-16cores  &  & 0.970  & & 465\\ 
GradSAT-BFGS-1core      &       & 0.954 & & 293\\ 
GradSAT-MMA-1core       &       & 0.950 & & 248\\ 
GradSAT-CG-1core        &       & 0.949 & & 234\\
GradSAT-SLSQP-1core        &       & 0.937 & & 206 \\ 
WalkSAT    &          & 0.917    &  & 124\\ 
FourierSAT    &        & 0.908   &    & 108\\ 
Mixing Method  &       & 0.892   &    & 121 \\ 
Loandra     &          & 0.874   &    & 41\\ 
VBS         &          & 1       &    & 575\\ 
VBS without GradSAT &  & 0.990   &    & 254\\ 
	\bottomrule
\end{tabular}
\caption{\textbf{Results on MaxSAT instances}.  \toolname{} implementations achieve better avg. score as well as \#Win and improve the VBS.}
\label{table:maxsat}
\end{table}

\textbf{RQ1.} Table \ref{table:grad_speedup} lists the running time for computing gradient on several instances. Each running time is the average  on 100 random points in $[-1,1]^n$. The results indicate that 
\toolname{} significantly accelerates the computation of gradient  compared with FourierSAT. 

\textbf{RQ2.} We classify the four types of benchmarks in Benchmark 1 into two groups: 1) PBs and CARDs; 2) XORs-1CARD and CNF-XORs; because solvers have similar performance on instances from the same group. We summarize the results in Figure \ref{exp:rand_card} and \ref{exp:rand_xor}. Figure \ref{exp:rand_card} shows the results on CARDs and PBs instances. Though CDPB-based solver RoundingSAT easily solves all instances, \toolname{} on 16 cores is able to solve 732 instances, more than all other CDCL, DLS SAT solvers and CDCL-based PB solvers. From Figure \ref{exp:rand_xor}, we observe that on benchmarks involving XORs, CDCL solvers, especially CMS, perform better than local-search solvers. \toolname{} on 16 cores solves 155 instances, only second to CCAnr (191) among all local search solvers. We  conjecture that the solution space shattering \cite{cnfxor-phase-transition,cnfxor-card-phase-transition} brought by XORs might be the reason for the weakness of local search solvers in handling XORs.

\textbf{RQ3.} Table \ref{table:maxsat} showcases the results on MaxSAT benchmarks. We use the incomplete score \cite{loandra} and the number of instances where a solver provides the solution with the best cost (\#Win) as the metric for evaluation. All five versions of \toolname{}  achieve better average score and \#Win than other solvers. Moreover, \toolname{} improves the score of Virtual Best Solver (VBS)  with 0.01 and \#Win with 321. 

We also try \toolname{} on large-size instances from the \texttt{industrial} MaxSAT problems, which contain both hard and soft clauses. But \toolname{} is not able to give an answer for about $2/3$ of all instances. On instances which \toolname{} successfully solves, an average score of 0.82 is achieved.  We conjecture two reasons for the unsatisfactory performance of \toolname{} on \texttt{industrial} benchmarks as follows. First, \toolname{} sets the weight of hard clauses to be (\#soft clauses + 1). It has been shown \cite{satlike} that this natural weighting performs poorly without exploiting the structure of partial MaxSAT problems. Second, industrial benchmarks contain a large number of variables, which also the non-stochastic DLS solver, e.g., GSAT,  could hardly handle. Thus it is not surprising that GradSAT, as a proof-of-concept implementation demonstrating how to calculate the full gradients simultaneously, is less competitive. We leave handling industrial partial MaxSAT problems by CLS approaches as a future direction.

\textbf{RQ4.} Table \ref{table:nodes-sharing} shows the total number of nodes of individual BDDs and the number of nodes of MRBDD for each type of benchmarks In order to allow more nodes sharing, we use the same variable order for all individual BDDS, corresponding to the original variable indices. We leave planning and applying a better variable order as a future direction.  For all tracks of MaxSAT instances in CNF format, using MRBDD significantly reduced the number of nodes, yielding a average reduction ratio from 2.29 to 3.22. However, on non-CNF hybrid SAT instances, nodes sharing in MRBDD is negligible, probably due to relatively large constraint length.

\section{Conclusion and Future Directions}
In this paper, we proposed a new BDD-based continuous local search (CLS) framework for solving Boolean formulas consisting of hybrid constraints, including symmetric constraints and pseudo-Boolean constraints. The core of our approach is a novel algorithm for computing the  gradient as fast as evaluating the objective function.  We implemented our method as \toolname{} on top of a portfolio of continuous optimizers. The experimental results demonstrate the power of \toolname{}  at handling  CARD/PB constraints and small-size MaxSAT problems, while we also observe the limitations of our tool on large instances and XOR-rich instances.

Since for large instances, the cost for computing the gradient is still too high, a future direction is applying stochastic gradient descent to reduce the computational cost. Another future direction is to combine the advantages of  CLS methods, i.e., the capability of ``searching inside" and those of  DLS methods, including ``sideway walk", which searches the space quickly, in order to design stronger solvers.  We also want to design better weighting mechanism 
as in \cite{satlike} to handle partial MaxSAT problems containing hard constraints.
    
\clearpage
\section*{Acknowledgement}
Work supported in part by NSF grants IIS-1527668, CCF-1704883,
IIS-1830549, DoD MURI grant N00014-20-1-2787, and an award from the Maryland Procurement Office.
\bibliographystyle{unsrt}  
\bibliography{ref} 

\begin{thebibliography}{10}

\bibitem{Vardi14a}
Moshe~Y. Vardi.
\newblock Boolean satisfiability: theory and engineering.
\newblock {\em Commun. {ACM}}, 57(3):5, 2014.

\bibitem{marques1999grasp}
Joao~P Marques-Silva and Karem~A Sakallah.
\newblock {{GRASP: A} Search Algorithm For Propositional Satisfiability}.
\newblock {\em IEEE Transactions on Computers}, 48(5):506--521, 1999.

\bibitem{davis1960computing}
Martin Davis and Hilary Putnam.
\newblock {A Computing Procedure for Quantification Theory}.
\newblock {\em Journal of the ACM (JACM)}, 7(3):201--215, 1960.

\bibitem{davis1962machine}
Martin Davis, George Logemann, and Donald Loveland.
\newblock {A Machine Program for Theorem-Proving}.
\newblock {\em Communications of the ACM}, 5(7):394--397, 1962.

\bibitem{chaff_paper}
Matthew~W. Moskewicz, Conor~F. Madigan, Ying Zhao, Lintao Zhang, and Sharad
  Malik.
\newblock Chaff: Engineering an efficient sat solver, 2001.

\bibitem{minisat}
Niklas E{\'e}n and Niklas S{\"o}rensson.
\newblock {An Extensible SAT-Solver}.
\newblock In {\em SAT}, pages 502--518, 2003.

\bibitem{PicoSAT}
Armin Biere.
\newblock {PicoSAT Essentials}.
\newblock {\em JSAT}, 4:75--97, 2008.

\bibitem{lingeling}
Armin Biere.
\newblock {Lingeling , Plingeling , PicoSAT and PrecoSAT at SAT Race 2010},
  2010.

\bibitem{glucose}
Gilles Audemard and Laurent Simon.
\newblock {Lazy Clause Exchange Policy for Parallel SAT Solvers}.
\newblock In {\em SAT 2014}, pages 197--205, Cham, 2014.

\bibitem{maplesat}
Jia~Hui Liang.
\newblock {\em Machine Learning for SAT Solvers}.
\newblock PhD thesis, University of Waterloo, December 2018.

\bibitem{GSAT}
Bart Selman, Hector Levesque, and David Mitchell.
\newblock {A New Method for Solving Hard Satisfiability Problems}, 1992.

\bibitem{walksat}
Bart Selman, Henry Kautz, and Bram Cohen.
\newblock {Local Search Strategies for Satisfiability Testing}.
\newblock {\em Second DIMACS Implementation Challenge}, 26, 09 1999.

\bibitem{Nsat}
David McAllester, Bart Selman, and Henry Kautz.
\newblock {Evidence for Invariants in Local Search}, 1997.

\bibitem{novelty}
Holger Hoos and Thomas Stützle.
\newblock Local search algorithms for sat: An empirical evaluation.
\newblock {\em J.~Automated Reasoning}, 24:421--481, 05 2000.

\bibitem{SAPS}
F.~Hutter, Dave A.~D. Tompkins, and H.~Hoos.
\newblock Scaling and probabilistic smoothing: Efficient dynamic local search
  for sat.
\newblock In {\em CP}, 2002.

\bibitem{probSAT}
Adrian Balint and Uwe Sch{\"o}ning.
\newblock {Choosing Probability Distributions for Stochastic Local Search and
  the Role of Make versus Break}.
\newblock In {\em SAT 2012}, pages 16--29, 2012.

\bibitem{ccanr}
Shaowei Cai, Chuan Luo, and Kaile Su.
\newblock Ccanr: A configuration checking based local search solver for
  non-random satisfiability.
\newblock In Marijn Heule and Sean Weaver, editors, {\em Theory and
  Applications of Satisfiability Testing -- SAT 2015}, pages 1--8, Cham, 2015.
  Springer International Publishing.

\bibitem{SATInteriorPoint}
A.~Kamath, N.~Karmarkar, K.~G. Ramakrishnan, and M.~C. Resende.
\newblock Computational experience with an interior point algorithm on the
  satisfiability problem.
\newblock {\em Annals of Operations Research}, 25:43--58, 1990.

\bibitem{continuouslocalsearch}
{Jun Gu}.
\newblock Global optimization for satisfiability (sat) problem.
\newblock {\em IEEE Transactions on Knowledge and Data Engineering},
  6(3):361--381, 1994.

\bibitem{Biclique-Cryptanalysis-of-the-Full-AES}
Andrey Bogdanov, Dmitry Khovratovich, and Christian Rechberger.
\newblock {Biclique Cryptanalysis of the Full AES}.
\newblock In {\em ASIACRYPT 2011}, 2011.

\bibitem{Graph-coloring-with-cardinality-constraints}
M.-C. Costa, D.~de~Werra, C.~Picouleau, and B.~Ries.
\newblock {Graph Coloring with Cardinality Constraints on the Neighborhoods}.
\newblock {\em Discrete Optimization}, 6(4):362 -- 369, 2009.

\bibitem{nae-coloring}
Irit Dinur, Oded Regev, and Clifford Smyth.
\newblock {The Hardness of 3-Uniform Hypergraph Coloring}.
\newblock {\em Combinatorica}, 25(5):519--535, Sep 2005.

\bibitem{encoding-handbook-of-satisfiability}
S.~Prestwich.
\newblock {CNF Encodings, Handbook of Satisfiability: Volume 185 Frontiers in
  Artificial Intelligence and Applications}, 2009.

\bibitem{Exploiting-Cardinality-Encodings-in-Parallel-Maximum-Satisfiability}
R.~{Martins}, V.~{Manquinho}, and I.~{Lynce}.
\newblock {Exploiting Cardinality Encodings in Parallel Maximum
  Satisfiability}.
\newblock In {\em ICTAI}, pages 313--320, Nov 2011.

\bibitem{cmspaper}
Mate Soos, Karsten Nohl, and Claude Castelluccia.
\newblock {Extending {SAT} Solvers to Cryptographic Problems}.
\newblock In {\em {SAT}}, pages 244--257, 2009.

\bibitem{Pueblo}
Hossein~M. Sheini and Karem~A. Sakallah.
\newblock Pueblo: A hybrid pseudo-boolean sat solver.
\newblock {\em JSAT}, 2:165--189, 2006.

\bibitem{zap}
H.~E. {Dixon}, M.~L. {Ginsberg}, E.~M. {Luks}, and A.~J. {Parkes}.
\newblock {Generalizing Boolean Satisfiability II: Theory}.
\newblock {\em arXiv e-prints}, page arXiv:1109.2134, September 2011.

\bibitem{fouriersat}
Anastasios {Kyrillidis}, Anshumali {Shrivastava}, Moshe~Y. {Vardi}, and Zhiwei
  {Zhang}.
\newblock {FourierSAT: A Fourier Expansion-Based Algebraic Framework for
  Solving Hybrid Boolean Constraints}.
\newblock In {\em AAAI 2020}, 2020.

\bibitem{modelCountingbyTensor}
Jeffrey~M. {Dudek} and Moshe~Y. {Vardi}.
\newblock {Parallel Weighted Model Counting with Tensor Networks}.
\newblock {\em arXiv e-prints}, page arXiv:2006.15512, June 2020.

\bibitem{countingIsHard}
Leslie~G. Valiant.
\newblock The complexity of enumeration and reliability problems.
\newblock {\em SIAM Journal on Computing}, 8(3):410--421, 1979.

\bibitem{BDD}
R.~E. {Bryant}.
\newblock Binary decision diagrams and beyond: enabling technologies for formal
  verification.
\newblock In {\em Proceedings of IEEE International Conference on Computer
  Aided Design (ICCAD)}, pages 236--243, 1995.

\bibitem{cudd}
Fabio Somenzi.
\newblock {CUDD: CU Decision Diagram Package Release 3.0.0}, 2015.

\bibitem{breakOut}
Paul Morris.
\newblock {The Breakout Method for Escaping from Local Minima}.
\newblock In {\em AAAI}, 1993.

\bibitem{DLM}
Benjamin Wah and Yi~Shang.
\newblock A discrete lagrangian-based global-search method for solving
  satisfiability problems.
\newblock {\em J Global Optimization}, 12(1), 07 1996.

\bibitem{Incomplete-algorithms}
Henry~A. Kautz, A.~Sabharwal, and B.~Selman.
\newblock Incomplete algorithms.
\newblock In {\em Handbook of Satisfiability}, 2009.

\bibitem{O'Donnell:2014:ABF:2683783}
Ryan O'Donnell.
\newblock {\em Analysis of Boolean Functions}.
\newblock Cambridge University Press, New York, NY, USA, 2014.

\bibitem{non-convex-is-nphard}
Prateek {Jain} and Purushottam {Kar}.
\newblock {Non-convex Optimization for Machine Learning}.
\newblock {\em arXiv e-prints}, page arXiv:1712.07897, December 2017.

\bibitem{nocedal2006numerical}
Jorge Nocedal and Stephen Wright.
\newblock {\em Numerical optimization}.
\newblock Springer Science \& Business Media, 2006.

\bibitem{larson2019derivative}
Jeffrey Larson, Matt Menickelly, and Stefan~M Wild.
\newblock Derivative-free optimization methods.
\newblock {\em Acta Numerica}, 28:287--404, 2019.

\bibitem{berahas2019theoretical}
Albert~S Berahas, Liyuan Cao, Krzysztof Choromanski, and Katya Scheinberg.
\newblock A theoretical and empirical comparison of gradient approximations in
  derivative-free optimization.
\newblock {\em arXiv preprint arXiv:1905.01332}, 2019.

\bibitem{gradient-free-is-slow}
Warren Hare, Julie Nutini, and Solomon Tesfamariam.
\newblock A survey of non-gradient optimization methods in structural
  engineering.
\newblock {\em Advances in Engineering Software}, 59:19 -- 28, 2013.

\bibitem{probabilityAssignment}
M.A. Thornton and V.S.S. Nair.
\newblock Efficient spectral coefficient calculation using circuit output
  probabilities.
\newblock {\em Digital Signal Processing}, 4(4):245 -- 254, 1994.

\bibitem{bddsat}
Guoqiang Pan and Moshe~Y. Vardi.
\newblock Symbolic techniques in satisfiability solving.
\newblock In Enrico Giunchiglia and Toby Walsh, editors, {\em SAT 2005}, pages
  25--50, Dordrecht, 2006. Springer Netherlands.

\bibitem{addmc}
Jeffrey~M. {Dudek}, Vu~H.~N. {Phan}, and Moshe~Y. {Vardi}.
\newblock {ADDMC: Weighted Model Counting with Algebraic Decision Diagrams}.
\newblock {\em arXiv e-prints}, page arXiv:1907.05000, July 2019.

\bibitem{PEARL1988143}
Judea Pearl.
\newblock Chapter 4 - belief updating by network propagation.
\newblock In Judea Pearl, editor, {\em Probabilistic Reasoning in Intelligent
  Systems}, pages 143 -- 237. Morgan Kaufmann, San Francisco (CA), 1988.

\bibitem{beliefPropogation}
Glenn Shafer and Prakash Shenoy.
\newblock Probability propagation.
\newblock {\em Ann Math Artif Intell}, 2:327--351, 03 1990.

\bibitem{symmetric-functions-BDD}
Tsutomu Sasao and Masahiro Fujita.
\newblock {\em Representations of Discrete Functions}.
\newblock Kluwer Academic Publishers, USA, 1996.

\bibitem{minisatplus}
Amir Aavani.
\newblock Translating pseudo-boolean constraints into cnf.
\newblock In Karem~A. Sakallah and Laurent Simon, editors, {\em Theory and
  Applications of Satisfiability Testing - SAT 2011}, pages 357--359, Berlin,
  Heidelberg, 2011. Springer Berlin Heidelberg.

\bibitem{slsqp2}
Dieter Kraft.
\newblock Algorithm 733: Tomp–fortran modules for optimal control
  calculations.
\newblock {\em ACM Trans. Math. Softw.}, 20(3):262–281, September 1994.

\bibitem{mma}
{Krister,Svanberg}.
\newblock A class of globally convergent optimization methods based on
  conservative convex separable approximations, 2002.

\bibitem{nlopt}
Steven~G. Johnson.
\newblock {The NLopt nonlinear-optimization package}, 2015.

\bibitem{cg}
Magnus~R. Hestenes and Eduard Stiefel.
\newblock Methods of conjugate gradients for solving linear systems.
\newblock {\em Journal of research of the National Bureau of Standards},
  49:409--435, 1952.

\bibitem{bfgs}
R.~Fletcher.
\newblock {\em Practical Methods of Optimization}.
\newblock Number v. 2 in A Wiley-Interscience publication. Wiley, 1987.

\bibitem{dlib09}
Davis~E. King.
\newblock Dlib-ml: A machine learning toolkit.
\newblock {\em Journal of Machine Learning Research}, 10:1755--1758, 2009.

\bibitem{Walksat-implementation}
Shaowei Cai, Kaile Su, and Chuan Luo.
\newblock Improving {W}alksat for {R}andom k-{S}atisfiability {P}roblem with k
  $>$ 3.
\newblock In {\em AAAI}, 2013.

\bibitem{openwbo}
Ruben Martins, Vasco Manquinho, and In{\^e}s Lynce.
\newblock Open-wbo: A modular maxsat solver,.
\newblock In Carsten Sinz and Uwe Egly, editors, {\em Theory and Applications
  of Satisfiability Testing -- SAT 2014}, pages 438--445, Cham, 2014. Springer
  International Publishing.

\bibitem{naps}
Masahiko Sakai and Hidetomo Nabeshima.
\newblock Construction of an robdd for a pb-constraint in band form and related
  techniques for pb-solvers.
\newblock {\em IEICE Transactions on Information and Systems},
  E98.D(6):1121--1127, 2015.

\bibitem{roundingsat}
Jan Elffers and Jakob Nordstr\"om.
\newblock {Divide and Conquer: Towards Faster Pseudo-Boolean Solving}, 7 2018.

\bibitem{loandra}
Jeremias Berg, Emir Demirovi{\'{c}}, and Peter~J. Stuckey.
\newblock Core-boosted linear search for incomplete maxsat.
\newblock In Louis-Martin Rousseau and Kostas Stergiou, editors, {\em
  Integration of Constraint Programming, Artificial Intelligence, and
  Operations Research}, pages 39--56, Cham, 2019. Springer International
  Publishing.

\bibitem{wang2017mixing}
Po-Wei Wang, Wei-Cheng Chang, and J.~Zico Kolter.
\newblock The mixing method: coordinate descent for low-rank semidefinite
  programming.
\newblock {\em arXiv preprint arXiv:1706.00476}, 2017.

\bibitem{pysat}
Alexey Ignatiev, Antonio Morgado, and Joao Marques{-}Silva.
\newblock {{PySAT:} {A} {Python} Toolkit for Prototyping with {SAT} Oracles}.
\newblock In {\em SAT}, pages 428--437, 2018.

\bibitem{xor-encoding}
Chu~Min Li.
\newblock {Integrating Equivalency Reasoning into Davis-Putnam Procedure}.
\newblock In {\em AAAI}, pages 291--296. AAAI Press, 2000.

\bibitem{cnfxor-phase-transition}
Jeffrey~M. {Dudek}, Kuldeep~S. {Meel}, and Moshe~Y. {Vardi}.
\newblock {Combining the $k$-CNF and XOR Phase-Transitions}.
\newblock In {\em IJCAI}, Feb 2016.

\bibitem{cnfxor-card-phase-transition}
Yash Pote, Saurabh Joshi, and Kuldeep~S. Meel.
\newblock {Phase Transition Behavior of Cardinality and XOR Constraints}.
\newblock In {\em {IJCAI-19}}, 7 2019.

\bibitem{satlike}
Zhendong Lei and Shaowei Cai.
\newblock Solving (weighted) partial maxsat by dynamic local search for sat.
\newblock In {\em Proceedings of the Twenty-Seventh International Joint
  Conference on Artificial Intelligence, {IJCAI-18}}, pages 1346--1352.
  International Joint Conferences on Artificial Intelligence Organization, 7
  2018.

\end{thebibliography}
\newpage
\section*{Appendix}
\label{appendix}
\subsection{Proof of Theorem \ref{theo:rounding}}
  We first prove that for the Walsh-Fourier Expansion of a clause $c$, denoted by $\texttt{WFE}$, the probability that $c$ is satisfied by $\mathcal{R}(a)$, i.e., $\mathop{\mathbb{P}}[\texttt{WFE}(\mathcal{R}(a))=1]$ equals $\texttt{WFE}(a)$. Then by the definition of the objective function and linearity of expectation, Theorem \ref{theo:rounding} follows directly.
        
        We  prove $\mathop{\mathbb{P}}[\texttt{WFE}(\mathcal{R}(a))=1]=\texttt{WFE}(a)$ by induction on the number of variables $n$.
        
		\noindent {\rm \texttt{Basis step}}: Let $n=1$. $\texttt{WFE}$ is either constant or $\frac{1+x_1}{2}$, or $\frac{1-x_1}{2}$. It is easy to verify that the statement holds.
		
		\noindent {\rm \texttt{Inductive step}}: Suppose $n\ge 2$. Then, by Boole's Expansion,  $\texttt{WFE}(\mathcal{R}(a))$ can be expanded as:
		
		\begin{align}\nonumber
		    \texttt{WFE}(\mathcal{R}(a))&=\tfrac{1-\mathcal{R}(a)_n}{2} \cdot  \texttt{WFE}_{n \to (-1)}(\mathcal{R}(a)_{[n-1]})
		  +\tfrac{1+\mathcal{R}(a)_{n}}{2} \cdot  \texttt{WFE}_{{n}\to 1}(\mathcal{R}(a)_{[n-1]}), 
		\end{align}
		where $\mathcal{R}(a)_n$ is the value of the $n$-th coordinate of $\mathcal{R}(a)$, $\mathcal{R}(a)_{[n-1]}\in \mathbb{R}^{n-1}$ is the point after removing the $n$-th coordinate of $\mathcal{R}(a)$, $\texttt{WFE}_{n\to 1}$ denotes the function of $\texttt{WFE}$ after fixing the value of $x_n$ to $1$.
		
		Note that the value of $\mathcal{R}(a)_n$ and $\mathcal{R}(a)_{[n-1]}$ are independent, thus
		\begin{equation}\nonumber
		\begin{split}
		    \mathop{\mathbb{P}}[ \texttt{WFE}(\mathcal{R}(a))=1]
		    =&\mathop{\mathbb{P}}[\mathcal{R}(a)_n=-1]\cdot \mathop{\mathbb{P}}[\texttt{WFE}_{n \to (-1)}(\mathcal{R}(a)_{[n-1]})=1]
		    + \mathbb{P}[\mathcal{R}(a)_n=1]\cdot \mathop{\mathbb{P}}[\texttt{WFE}_{n \to 1}(\mathcal{R}(a)_{[n-1]})=1]\\
		    =&\frac{1-a_n}{2}\cdot \texttt{WFE}_{n\to -1}(a_{[n-1]}) 
		    + \frac{1+a_n}{2}\cdot \texttt{WFE}_{n\to 1}(a_{[n-1]}) \text{ (by I.H.)} \\
		    =& \texttt{WFE}(a)
		\end{split} 
			\end{equation}
		\hfill\qedsymbol

\subsection{Proof of Theorem \ref{theo:main_weighted}}
Note that by $\mathop{\mathbb{P}}[\texttt{WFE}_c(\mathcal{R}(a))=1]=\texttt{WFE}_c(a)$ for all constraint $c\in C_f$, proved in the proof of Theorem \ref{theo:rounding},  we have $\texttt{WFE}_c(a)\in[0,1]$ for all constraint $c$ and $a\in[-1,1]^n$. Thus $F_{f,w}(a)\le \sum_{c\in C_f}w(c)$ for all $a\in [-1,1]^n$ since $w:C_F\to \mathbb{R}^+$ is a positive function.
		\begin{itemize}
			\item[--] "$\Rightarrow$":  Suppose $f$ is satisfiable and $b\in\{\pm 1\}^n$ is one of its solutions. Then $b$ is also a solution of every constraint of $f$. Thus for every $c\in C_f$, $\texttt{WFE}_c(b)=1$. Therefore $F_{f,w}(b)=\sum_{c\in C_f}w(c)$ and $\mathop{\max}\limits_{x\in [-1,1]^n}F_{f,w}(x)=\sum_{c\in C_f}w(c)$. 
			\item[--] "$\Leftarrow$": Suppose $\mathop{\max}\limits_{x\in [-1,1]^n}F_{f,w}(x)=\sum_{c\in C_f}w(c)$. Thus $\exists\, a$ such that $F_f(a)=\sum_{c\in C_f}w(c)$. Since $\texttt{WFE}_c(a)\in[0,1]$, we have $\texttt{WFE}_c(a)=1$ for every $c\in C_f$. Since $\texttt{WFE}_c(a) = \mathop{\mathbb{P}}[\texttt{WFE}_c(\mathcal{R}(a))=1]=1$, rounding $a$ arbitrarily to $b\in\{-1,1\}^n$ will give us $\texttt{WFE}_c(b)=1$ for all constraint $c$. Thus $b$ is a solution of all constraints, which is also a solution of the formula $f$. \hfill\qedsymbol
		\end{itemize}

\subsection{Proof of Corollary \ref{coro:cop}}
\begin{equation}\nonumber
\begin{split}
    \mathop{\mathbb{E}}\limits_{b\sim \mathcal{S}_a}[F_{f,w}(b)]
    = &\mathop{\mathbb{E}}\limits_{b\sim \mathcal{S}_a}[\sum\limits_{c\in C_f}w(c)\cdot \texttt{WFE}_c(b)]\\
      = &\sum\limits_{c\in C_f}w(c)\cdot \mathop{\mathbb{E}}\limits_{b\sim \mathcal{S}_a}[\texttt{WFE}_c(b)]\\
           = &\sum\limits_{c\in C_f}w(c)\cdot \mathop{\mathbb{P}}\limits_{b\sim \mathcal{S}_a}[c(b)=1]\\
             = &\sum\limits_{c\in C_f}w(c)\cdot \texttt{COP}(P_a,c)
\end{split}
\end{equation}
\subsection{Proof of Theorem \ref{lemma:forward}}

\begin{pfs*}
We first prove a result for single-rooted BDDs in Lemma \ref{lemma:singleBDDforward}. Then we show that running Algorithm \ref{algo:wmc} on individual single-rooted BDDs gives the same result with running Algorithm \ref{algo:wmc} on a corresponding multi-rooted BDD in Lemma \ref{lemma:mrbddforward}.
\end{pfs*}

\begin{lemma}
Let $B$ be a single-rooted BDD and run Algorithm \ref{algo:wmc} on $B$, a real assignment $a\in [-1,1]^n$ and the constraint weight $w$. Let $b\in \{-1,1\}^n$ be the randomly rounded assignment from $a$ by $\mathbb{P}[b_i=-1]=\frac{1-a_i}{2}$. Then for each node $v\in B.V$, we have 
$$
M_{TD}[v] = w \cdot \mathcal{P}(B,a,v),
$$
where $\mathcal{P}(B,a,v)$ is the probability that the node $v$ is on the path generated by $b$ on $B$.

Especially, 
$$
M_{TD}[B.one] = w \cdot \texttt{COP}(P_a,c)
$$
where $P_a(x_i)=\frac{1-a_i}{2}$ for all $i\in\{1,\cdots,n\}$.
\label{lemma:singleBDDforward}
\end{lemma}
\begin{proof*}
We prove the statement by structural induction on single-rooted BDD $B$.

\noindent {\rm \texttt{Basis step}}: For the root $r$ of $B$, $$M_{TD}[B.r] = 1$$ the statement holds because $r$ is on the path generated by every discrete assignment on $B$.

\noindent {\rm \texttt{Inductive step}}: For each non-root node $v$ of $B$, let $par_T(v,B)$ be the set of parents of $v$ with an edge labeled by \texttt{True} ($-1$) and $par_F(v,B)$ be the set of parents of $v$ with an edge labeled by \texttt{False} ($1$) in BDD $B$, after Algorithm \ref{algo:wmc} terminates, we have
\begin{equation} \nonumber
\begin{split}
M_{TD}[v] = \sum_{u\in par_T(v)}\frac{1-a_i}{2}\cdot{M_{TD}[u]} + \sum_{u\in par_F(v)}\frac{1+a_i}{2}\cdot{M_{TD}[u]}
    \end{split}
\end{equation}
By inductive hypothesis,
\begin{equation}\nonumber
\begin{split}
    M_{TD}[v] &= \sum_{u\in par_T(v)}w\cdot Pr[b_i=-1]\cdot \mathcal{P}(B,a,u)
    + \sum_{u\in par_F(v)}w\cdot Pr[b_i=1]\cdot\mathcal{P}(B,a,u)\\
    &= w\cdot\mathcal{P}(B,a,v)
    \end{split}
\end{equation}
The last equality holds because the events of reaching $v$ from different parents are exclusive.
\hfill\qedsymbol
\end{proof*}

Next we will generalize our result for multi-rooted BDDs.

\begin{lemma}
Let $C_f=\{c_1,\cdots, c_m\}$ be the constraints set of formula $f$. Let $B_1,\cdots,B_m$ be the individual BDDs corresponding to constraints in $C_f$. Let $B$ be the multi-rooted BDD for the whole set $C_f$.  Suppose we run:
\begin{itemize}
    \item Algorithm \ref{algo:wmc} with $(B,a,w)$ and store $M_{TD}$.
    \item Algorithm \ref{algo:wmc} with $(B_i,a,w(c_i))$ for each constraint $c_i$ and store $m$ mappings $\{M_{TD}^1,\cdots M_{TD}^m\}$.
    Then for each node $v\in B.V$, suppose the equivalent nodes of $v$ also appear in $\mathcal{B}_v = (B_{I_{v1}},\cdots,B_{I_{vk}})$ as $\mathcal{V}_v = (v_{J_{v1}},\cdots,v_{J_{vk}})$, respectively.
    \end{itemize}
    Then we have 
    $$
    M_{TD}[v] = \sum_{j=1}^{|I_v|}M_{TD}^{I_{vj}}[v_{J_{vj}}]
    $$
\label{lemma:mrbddforward}
\end{lemma}
\begin{proof*}
We prove by induction on the multi-rooted BDD $B$.

\noindent {\rm \texttt{Basis step}}: For each node $r\in B.V$ with in-degree 0, we have 

$$M_{TD}[v] =\sum_{j=1}^{{|I_v|}} w(c_{I_{vj}})= \sum_{j=1}^{{|I_v|}}M_{TD}^{I_{vj}}[v_{J_{vj}}],$$
since $v$ does not have parent nodes in $B$ and all $B_i$'s.

\noindent {\rm \texttt{Inductive step}}:
For each node $v\in B.V$ that has at least one parent, let $par_T(v)$ be the set of parents of $v$ with an edge labeled by \texttt{True} ($-1$) and $par_F(v)$ be the set of parents of $v$ with an edge labeled by \texttt{False} ($1$). The equivalence of $v$ can be either a root in some BDDs or non-root nodes in some other BDDs. Suppose the equivalence of $v$ is the root in BDDs with constraint index $R_v = \{R_{v1},\cdots,R_{vd}\}$.

Consider the situation after  Algorithm \ref{algo:wmc} terminates on MRBDD $B$, we have

\begin{equation} \nonumber
\begin{split}
M_{TD}[v] &= \sum_{u\in par_T(v,B)}\frac{1-a_i}{2}\cdot{M_{TD}[u]} + \sum_{u\in par_F(v,B)}\frac{1+a_i}{2}\cdot{M_{TD}[u]}+ \sum_{j=1}^{|R_v|} w(c_{vj})
    \end{split}
\end{equation}
By inductive hypothesis, 

$$M_{TD}[u] = \sum_{j=1}^{{|I_u|}}M_{TD}^{I_{uj}}[v_{J_{uj}}].$$
for all $u\in par_T(v,B)\cup par_F(v,B)$. 

Thus 
\begin{equation} \nonumber
\begin{split}
M_{TD}[v] &= \sum_{u\in par_T(v,B)}\frac{1-a_i}{2}\cdot \sum_{j=1}^{|I_u|}M_{TD}^{I_{uj}}[v_{J_{uj}}] + \sum_{u\in par_F(v,B)}\frac{1+a_i}{2}\cdot\sum_{j=1}^{|I_u|}M_{TD}^{I_{uj}}[v_{J_{uj}}]+ \sum_{j=1}^{|R_v|} w(c_{R_{vj}}) \\
&= \sum_{j=1}^{|I_v|}\Big(\sum_{u\in par_T(v,B_{I_{vj}})}\frac{1-a_i}{2}\cdot M_{TD}^{I_{uj}}[v_{I_{uj}}] + \sum_{u\in par_F(v,B_{I_{vj}})}\frac{1+a_i}{2}\cdot M_{TD}^{I_{uj}}[v_{I_{uj}}] + \mathbb{I}({I_{vj}}\in R_v) \cdot w(c_{R_{vj}}) \Big) \\
&=\sum_{j=1}^{|I_v|}M_{TD}^{I_{vj}}[v_{I_{vj}}]
    \end{split}
\end{equation}
\hfill\qedsymbol
\end{proof*}
Now we are ready to prove Theorem \ref{lemma:forward}.

By the fact that the node labeled by constant \texttt{one} appears in every individual BDD $B_i$, we have 
\begin{equation}\nonumber
\begin{split}
    M_{TD}[B.one] &=  \sum_{c_i\in C_f}M_{TD}^i[B_i.one]  \text{\,\,\,\,(Lemma \ref{lemma:mrbddforward})}\\
   &= \sum_{c_i\in C_f}w(c_i)\cdot \texttt{COP}(P_a,c_i) \text{\,\,\,\,(Lemma \ref{lemma:singleBDDforward})}\\
   &= F_{f,w}(a). \text{\,\,\,\,(Corollary \ref{coro:cop})}
\end{split}
\end{equation}

\textbf{Complexity}: Algorithm \ref{algo:wmc} traverse the multi-rooted BDD once. The topological sort also takes $O(S)$ time (e.g. Kahn's algorithm). In practice, since the topological order is indicated by the variable index, topological sort can be conveniently implemented by a priority queue with the variable index as the key, although this would make the complexity $O(SlogS)$.  \hfill\qedsymbol
\appendix

\subsection{Proof of Theorem \ref{lemma:backward}}
To prove Theorem \ref{lemma:backward}, we introduce the following lemma.
\begin{lemma}
Suppose Algorithm \ref{algo:wmc-bottomup} with inputs $(B,a,w)$ terminates, where $B$ is a multi-rooted BDD. Then for each node $v\in B.V$, let $f_v$ be the sub-function corresponding to the sub-BDD generated by regarding $v$ as the root. The following holds:
$$
M_{BU}[v] = \texttt{COP}(P_a,f_v),
$$
where $P_a(x_i)=\frac{1-a_i}{2}$ for all $i\in \{1,\cdots,n\}$.
\label{lemma:backwardsingle}
\end{lemma}
\begin{proof*}
We prove by structural induction on multi-rooted BDD $B$. 

\noindent {\rm \texttt{Basis step}}: For the node $B.one$, $M_{BU}[v]=1$.  For the node $B.zero$, $M_{BU}[v]=0$. Since the sub-functions given by $B.one$ and $B.zero$ are the constant functions with value $1$ and $0$ respectively, the statement holds.

\noindent {\rm \texttt{Inductive step}}: For each non-terminal node $v$, after Algorithm \ref{algo:wmc-bottomup} terminates, we have

$$
M_{BU}[v] = \frac{1-a_i}{2}\cdot M_{BU}[v.T] + \frac{1+a_i}{2} \cdot M_{BU}[v.F].
$$

By inductive hypothesis, we have
\begin{equation}\nonumber
\begin{split}
M_{BU}[v] &= \frac{1-a_i}{2}\cdot \texttt{COP}(P_a,f_{v.T}) + \frac{1+a_i}{2} \cdot \texttt{COP}(P_a,f_{v.F})    \\
&=P(x_i)\cdot \texttt{COP}(P_a,f_{v.T}) + (1-P(x_i)) \cdot \texttt{COP}(P_a,f_{v.F}) \\
&=\texttt{COP}(P_a,f_{v})
\end{split}
\end{equation} \qedsymbol
\end{proof*}
Now we are ready to prove Theorem \ref{lemma:backward}. When Algorithm \ref{algo:wmc-bottomup} terminates, we have

\begin{equation} \nonumber
    \begin{split}
        &\sum_{c\in C_f}(M_{BU}[B.entry(c)]\cdot w(c))\\
        &= \sum_{c\in C_f} \texttt{COP}(P_a,f_{B.entry(c)})\cdot w(c) \text{\,\,\,\, (Lemma \ref{lemma:backwardsingle})}\\
         &= \sum_{c\in C_f} \texttt{COP}(P_a,c)\cdot w(c) \\
         &=F_{f,w}(a) \text{\,\,\,\, (Corollary \ref{coro:cop})}
    \end{split}
\end{equation}

\textbf{Complexity}: The analysis is similar with the proof of Theorem \ref{algo:wmc}. 
\hfill\qedsymbol

\subsection{Proof of Theorem \ref{theo:gradient_is_fast}}
\begin{pfs*}
In the proof of Theorem \ref{lemma:forward} and \ref{lemma:backward} we actually reveal the meaning of the values of mappings $M_{TD}$ and $M_{BU}$ on a node $v$. Roughly speaking, $M_{TD}[v]$ has connection with the probability of a BDD node $v$ being reached from the root ($\mathcal{P}(B,a,v)$), while $M_{BU}[v]$ contains information of the circuit-output probability of the sub-function defined by the sub-BDD with  root $v$ (\texttt{COP}$(P_a,f_v)$). Intuitively, Algorithm \ref{algo:grad} can be interpreted as applying the differentiation operation on each node. We prove this theorem from writing down the definition of gradient and by gradually massaging the representation of gradient  to connect it with $M_{TD}$ and $M_{BU}$.
\end{pfs*}

By Corollary \ref{coro:cop}, for a real point $a\in [-1,1]^n$, we have

\begin{equation} \nonumber
    \begin{split}
        F_{f,w}(a) &=\sum_{c\in C_f}w(c)\cdot \texttt{COP}(P_a,c)\\
    \end{split}
\end{equation}
Thus the gradient of $F_{f,w}$ for $x_i$ at a real point $a$, denoted by $\frac{\partial F_{f,w}}{\partial x_i}(a)$, can be computed by:
\begin{equation} \nonumber
    \begin{split}
       \frac{\partial F_{f,w}}{\partial x_i}(a)&=\sum_{c\in C_f}w(c)\cdot \big( \texttt{COP}(P_{{a_{i+}}},c) 
       - \texttt{COP}(P_{{a_{i-}}},c) \big),\\
    \end{split}
\end{equation}
where $a_{i+}=(a_1,\cdots,a_i=1,\cdots, a_n)$ and $a_{i-}=(a_1,\cdots,a_i=-1,\cdots, a_n)$.
Let $X_i=(x_1,\cdots,x_{i-1},x_{i+1},\cdots,x_n)$, $X_{i,1}=(x_1,\cdots, x_{i-1})$ and $X_{i,2}=(x_{i+1},\cdots, x_n)$. 

By the definition of \texttt{COP} we have
\begin{equation} \nonumber
\hspace{-2cm}
    \begin{split}
    &\texttt{COP}(P_{a_{i+}},c)\\
 =&\sum_{b\in \{-1,1\}^{n-1}}c(X_i=b,x_i=1) \cdot \prod_{b_j=-1} P_a(x_j) \cdot \prod_{b_j=1}(1-P_a(x_j))\\
=&\sum_{b_1\in \{-1,1\}^{i-1}}  \Big(\prod_{\substack{b_{1j}=-1\\ x_j\in X_{i,1}}} P_a(x_j)\cdot \prod_{\substack{b_{1j}=1\\ x_j\in X_{i,1}}}(1-P_a(x_j)) \Big)\cdot
\sum_{b_2\in \{-1,1\}^{n-i}}  \Big(\prod_{\substack{b_{2j}=-1\\ x_{i+j}\in X_2}} P_a(x_{i+j})\cdot \prod_{\substack{b_{2j}=1\\ x_{i+j}\in X_2}}(1-P_a(x_{i+j}))\cdot c(b_1,1,b_2) \Big)\\
\end{split}
\end{equation}
Thus 
\begin{equation} \nonumber
    \begin{split}
     &\frac{\partial F_{f,w}}{\partial x_i}(a)\\ =&\sum_{c\in C_f}w(c)\cdot\big(\texttt{COP}(P_{{a_{i+}}},c) - \texttt{COP}(P_{{a_{i-}}},c)\big) \\
     =& \sum_{c\in C_f}w(c) \cdot \sum_{b_1\in \{-1,1\}^{i-1}}  \Big(\prod_{\substack{b_{1j}=-1\\ x_j\in X_1}} P_a(x_j)\cdot \prod_{\substack{b_{1j}=1\\ x_j\in X_1}}(1-P_a(x_j)) \Big)\cdot
\Big(\sum_{b_2\in \{-1,1\}^{n-i}}  \prod_{\substack{b_{2j}=-1\\ x_{i+j}\in X_2}} P_a(x_{i+j})\cdot \prod_{\substack{b_{2j}=1\\ x_{i+j}\in X_2}}(1-P_a(x_{i+j}))\cdot \\
&\big(c(b_1,1,b_2)-c(b_1,-1,b_2) \Big)\\
    \end{split}
\end{equation}

Note that the following part of the right hand side of the equation above,
\begin{equation}\nonumber
    \begin{split}
&\Big(\sum_{b_2\in \{-1,1\}^{n-i}}  \prod_{\substack{b_{2j}=-1\\ x_{i+j}\in X_2}} P_a(x_{i+j})\cdot \prod_{\substack{b_{2j}=1\\ x_{i+j}\in X_2}}(1-P_a(x_{i+j}))\cdot \big(c(b_1,1,b_2)-c(b_1,-1,b_2) \Big)
    \end{split}
\end{equation}
is non-zero if and  only if the value of $b_1=(b_{11},\cdots,b_{1,i-1})$ leads to a non-terminal node of BDD $B_c$ labeled by $x_i$, where $B_c$ is the BDD corresponding with constraint $c$.

Thus we are able to simplify the representation of gradient by getting rid of $b_1$ and the equation above can be written as

\begin{equation} \nonumber
\hspace{-2cm}
    \begin{split}
     &\frac{\partial F_{f,w}}{\partial x_i}(a)\\ 
     =& \sum_{c\in C_f}w(c)\sum_{\substack{v\in B_c\\ i_v=i}} \mathcal{P}(B_c,a,v)\cdot
\Big(\sum_{b_2\in \{-1,1\}^{n-i}}  \prod_{\substack{b_{2j}=-1\\ x_{i+j}\in X_2}} P_a(x_{i+j})\cdot \prod_{\substack{b_{2j}=1\\ x_{i+j}\in X_2}}(1-P_a(x_{i+j}))\cdot 
\big(f_v(x_i=1,X_{i,2}=b_2)-f_v(x_i=-1,X_{i,2}=b_2) \Big)\\
    \end{split}
\end{equation}

Note that the term 
\begin{equation}\nonumber
\begin{split}
    &\sum_{b_2\in \{-1,1\}^{n-i}}  \prod_{\substack{b_{2j}=-1\\ x_{i+j}\in X_2}} P_a(x_{i+j})\cdot \prod_{\substack{b_{2j}=1\\ x_{i+j}\in X_2}}(1-P_a(x_{i+j}))\cdot f_v(x_i=-1,X_{i,2}=b_2)
\end{split}
\end{equation}
equals $\texttt{COP}({P_a},v.T)$.

Similarly,  
\begin{equation}\nonumber
\begin{split}
    &\sum_{b_2\in \{-1,1\}^{n-i}}  \prod_{\substack{b_{2j}=-1\\ x_{i+j}\in X_2}} P_a(x_{i+j}) \cdot \prod_{\substack{b_{2j}=1\\ x_{i+j}\in X_2}}(1-P_a(x_{i+j}))  \cdot f_v(x_i=1,X_{2,i}=b_2)
\end{split}
\end{equation}
 equals $\texttt{COP}({P_a},v.F)$.

Thus we further simplify the equation for computing gradient:
\begin{equation} \nonumber
    \begin{split}
     &\frac{\partial F_{f,w}}{\partial x_i}(a)
     = \sum_{c_i\in C_f}\sum_{\substack{v\in B_i\\ i_v=i}} w(c)\mathcal{P}(B_i,a,v)
     \cdot
\Big(\texttt{COP}(P_a,v.F)-\texttt{COP}(P_a,v.T)\Big)\\
    \end{split}
\end{equation}
Next we need to involve results given by Algorithm \ref{algo:grad} and the multi-rooted BDD $B$ instead of using separate BDDs for each constraints. Suppose the call to Algorithm \ref{algo:wmc} and \ref{algo:wmc-bottomup} in line 3-4 of Algorithm \ref{algo:grad} terminate. Then we have,
\begin{equation}\nonumber
\begin{split}
    &\sum_{c_i\in C_f}\sum_{\substack{v\in B_i\\ i_v=i}} w(c)\mathcal{P}(B_i,a,v) \\
    =&\sum_{c_i\in C_f}\sum_{\substack{v\in B_i\\ i_v=i}} M^i_{TD}[v]\text{\,\,\,\,\,\,(Lemma \ref{lemma:singleBDDforward})}\\
    =&\sum_{\substack{v\in B\\ i_v=i}}  M_{TD}[v]
\end{split}    
\end{equation}

On the other hand, note that $\texttt{COP}(P_a,v)$ remains the same despite of whether $v$ is in an individual BDD of a specific constraint or in the multi-rooted BDD.
 Thus,
\begin{equation} \nonumber
    \begin{split}
     &\frac{\partial F_{f,w}}{\partial x_i}(a)\\ 
     =& \sum_{\substack{v\in B\\ i_v=i}} M_{TD}[v]\cdot
\Big(\texttt{COP}(P_a,v.F)-\texttt{COP}(P_a,v.T)\Big)\\
=&\sum_{\substack{v\in B\\ i_v=i}} M_{TD}[v]\cdot \big(M_{BU}[v.F]-M_{BU}[v.T]\big) \text{\,\,\,\, (Lemma \ref{lemma:backwardsingle})}
    \end{split}
\end{equation}
The right hand side of the equation above is exactly what line 5-7 of Algorithm \ref{algo:grad} computes.

\textbf{Complexity:} Algorithm \ref{algo:grad} invokes Algorithm \ref{algo:wmc} and \ref{algo:wmc-bottomup}, with complexity both $O(S)$ by Theorem \ref{lemma:forward} and \ref{lemma:backward}. Line 5-8 of Algorithm \ref{algo:grad} can also be done in $O(S)$. In fact, line 5-8 of Algorithm \ref{algo:grad} can be combined with Algorithm \ref{algo:wmc-bottomup} once the top-down traverse is done so that the MRBDD is traversed twice instead of three times. Thus the complexity of  Algorithm \ref{algo:grad} is $O(S)$.

\hfill
\qedsymbol

\subsection{Generation of Benchmark 1}
In the following we will give a detailed description of how we generate the random hybrid Boolean benchmarks used in Section \ref{sec:exp} as \textbf{Benchmark 1}. All the instances generated are satisfiable. Let the number of variables be $n$, the CNF density (number of disjunctive clauses$/n$) be $r_C$, the XOR  density (number of XOR constraints$/n$) be $r_X$, the density of pseudo Boolean/cardinality constraints be $r_P$, the threshold of pseudo-Boolean constraints and cardinality constraints be $\delta$, let the density of variables in a cardinality/Pseudo-Boolean be $r_V$.  We generate the following four families of benchmarks. All benchmarks can be found in the .zip file.

\textbf{CNF-XORs.} We follow the setting of generating formulas with random 3CNF-XOR and 5-CNF-XOR in \cite{cnfxor-phase-transition}, where the phase-transition of CNF-XOR formulas is studied. For each variable number $n\in\{50,100,150\}$, for $(r_C, r_X)\in\{(1,0.2),(2,0.2),(3,0.2),(1,0.4),(2,0.4),(1,0.6)\}$, generate 10 random instance with $r_C\cdot n$ 3-CNF constraints, $r_X\cdot n$ XOR instances with the probability of each variable appears in a constraint $\frac{1}{2}$. For each variable number $n\in\{50,100,150\}$, for $(r_C, r_X)\in\{(5,0.2),(5,0.4),(5,0.6),(10,0.2),(10,0.4),(15,0.2)\}$, generate 10 random instance with $r_C\cdot n$ 3-CNF constraints, $r_X\cdot n$ XOR constraints with the probability of each variable appears in a constraint $\frac{1}{2}$. Total number of instances in this family: 360.

\textbf{XORs-1CARD.} We follow the setting of generating random XOR with 1 cardinality constraint in \cite{cnfxor-card-phase-transition} where the phase-transition of random XOR + 1 cardinality constraint is studied. For each variable number $n\in\{50,100,150\}$, $r_X\in\{0.2,0.3,0.4\}$, $\delta\in\{0.2,0.3,0.4\}$, we generate 10 random instance with $r_X\cdot n$ XOR constraints with the probability of each variable appears in a constraint $\frac{1}{2}$, as well as 1 global cardinality constraint $\sum_{i=1}^nx_i\le \delta\cdot n$. Total number of instances in this family: 270.

\textbf{CARDs.} Instances in this family are composed of random cardinality constraints. For each variable number $n\in\{50,100,150\}$, $r_P\in\{0.5,0.6,0.7\}$, $r_V\in\{0.2,0.3,0.4,0.5\}$, we generate 10 instances with each consisting $r_P\cdot n$ cardinality constraints. Each cardinality constraint contains  $r_V\cdot n$ variables randomly sampled from all $n$ variables. The direction is either "$\ge$" or "$\le$" with probability $\frac{1}{2}$ and the right-hand-side threshold is set to be $\frac{r_V\cdot n}{2}$. Total number of instances in this family: 360.

\textbf{PBs.} Instances in this family are composed of random pseudo-Boolean constraints. We generate two types of PBs instances according to how the coefficients of a PB constraint are obtained. 

1) For each appearance of a variable $x_i$ in a constraint, we sample an integer from $\{1,\cdots,n\}$ uniformly at random to be the coefficient of $x_i$. The coefficients of the same variable can be different in different constraints. For each variable number $n\in\{50,100,150\}$, $r_P\in\{0.5,0.6,0.7\}$, $r_V\in\{0.2,0.3,0.4,0.5\}$, we generate 10 instances with each consisting $r_P\cdot n$ PB constraints. Each PB constraint contains  $r_V\cdot n$ variables sampled uniformly at random from all $n$ variables. The direction is either "$\ge$" or "$\le$" with probability $\frac{1}{2}$. The right-hand-side threshold is set to be half the sum of coefficients in the left-hand-side.

2) For each variable $x_i$, we sample an integer from $\{1,\cdots,n\}$ uniformly at random to be the coefficient of $x_i$. The coefficient of the same variable will be the same in different constraints. For each variable number $n\in\{50,100,150\}$, $r_P\in\{0.5,0.6,0.7\}$, $r_V\in\{0.2,0.3,0.4,0.5\}$, we generate 10 instances with each consisting $r_P\cdot n$ PB constraints. Each PB constraint contains  $r_V\cdot n$ variables sampled uniformly at random from all $n$ variables. The direction is either "$\ge$" or "$\le$" with probability $\frac{1}{2}$. The right-hand-side threshold is set to be half the sum of coefficients in the left-hand-side.

 Total number of instances in this family: 720.

\end{document}